\documentclass[10pt]{article}
\usepackage{graphicx}
\usepackage[final]{colm2026_conference}
\usepackage{latexsym}
\usepackage{longtable}
\usepackage{microtype}
\usepackage{url}
\usepackage[dvipsnames]{xcolor}
\usepackage{enumitem}
\usepackage{xparse}
\usepackage{fancyvrb}
\usepackage{subcaption}
\usepackage{adjustbox}
\usepackage{booktabs}
\usepackage{float}
\usepackage{wrapfig}
\usepackage{tabularx}
\usepackage{array}
\usepackage{pifont}
\usepackage{colortbl}
\usepackage{hhline}
\usepackage[most]{tcolorbox}
\usepackage{bold-extra}
\usepackage{makecell}
\usepackage{tcolorbox}
\usepackage{hyperref}

\hypersetup{colorlinks, linkcolor={red!55!black}, citecolor={blue!65!black}, urlcolor={blue!65!black}}

\newcommand{\rOne}{\color{black}}

\let\realItem\item 
\makeatletter
\NewDocumentCommand\myItem{ o }{%
   \IfNoValueTF{#1}%
      {\realItem}
      {\realItem[#1]\def\@currentlabel{#1}}
}
\makeatother

\title{Language Models Might Not Understand You: \\ Evaluating Theory of Mind via Story Prompting}
\author{Nathaniel Getachew \& Abulhair Saparov \\
\texttt{\{ngetach, asaparov\}@purdue.edu} \\
}
\date{January 2026}

\begin{document}

\maketitle

\begin{abstract}
    We introduce \texttt{StorySim}, a programmable framework for synthetically generating stories to evaluate the \emph{theory of mind} (ToM) and \emph{world modeling} (WM) capabilities of large language models (LLMs). Unlike prior benchmarks that may suffer from contamination in pretraining data, or rely on an LLM for generation, \texttt{StorySim} produces novel, compositional story prompts anchored by a highly controllable \textsc{Storyboard}, enabling precise manipulation of character perspectives and events. Using \texttt{StorySim}, we evaluate LLMs across three kinds of ToM tasks: false belief, goal-directed, and preference attribution tasks. We then use our framework to design first- and second-order ToM tasks alongside WM tasks that control for the ability to track and model mental states. Our experiments across a suite of LLMs show that most models achieve higher accuracy on WM tasks than on ToM tasks, and that some models tend to reason more accurately when the subject of reasoning is a person rather than an inanimate object. Additionally, models struggle with goal-directed reasoning the most, and performance does not always scale with model size. All code for generating data and evaluations is freely available.\footnote{Code for and data generated by \texttt{StorySim} is available here: \href{https://github.com/ngetachew/StorySim}{\texttt{github.com/ngetachew/StorySim}}}

\end{abstract}

\section{Introduction} \label{sec:intro}

Large language models (LLMs), especially models meant for reasoning, have shown increasing improvement on benchmarks that were once infamously challenging. Contemporary models have made progress in reasoning through math problems, coding, and pure logical deduction \citep{jaech2024openai, guo2025deepseek, xia2025mimo}. As reasoning capabilities of LLMs continue to improve, we focus on LLMs’ ability to reason about mental states and the perspectives of other intelligent agents, a concept known as \emph{theory of mind} (ToM). This has become increasingly important, as an LLM without a sound ToM might not be able to understand the perspective of a user with harmful intent, or consider the effect its decisions could have on other people, who have their own goals and wishes. Reasoning about the perspective of other agents requires understanding how those agents' views are affected by changes in the environment. The ability to learn and adhere to these environmental rules is called world modeling (WM) \citep{bower1990mental, ha2018world, vafa2024evaluating}.  In principle, agents with strong WM should exhibit ToM, since other minds exist within the same world. 

ToM is often assessed in children by evaluating their ability to understand that others hold mental representations of the world that differ from their own. The prototypical ToM test is known as the \textit{Sally-Anne} test \citep{baron1985does}, where children are given a story akin to the following:
\begin{tcolorbox}[
colback=gray!15,
colframe=black!70,
arc=3pt,
boxrule=0.6pt,
left=3pt,right=3pt,
top=1.5pt,bottom=1.5pt,
title=\textbf{\emph{Classic Sally-Anne story}}
]
    \textcolor{RoyalBlue}{\small \textit{Sally and Anne walk into a bedroom. Sally puts a toy on top of the bed and leaves. Anne takes the toy and hides it under the bed. When Sally comes back into the room, where is the first place she will look for the toy?}}
\end{tcolorbox}

It may seem obvious to adults that the first place Sally will look is on top of the bed. However, it has been shown that children under the age of 4.5 think that Sally will look under the bed \citep{memisevic2018exploring}. This is hypothesized to be due to the fact that they have not yet developed ToM, so they cannot understand the fact that Sally and Anne have a different understanding of the toy's location.

\begin{figure*}
    \centering
    \includegraphics[trim={8em 7.5em 8em 0},clip,width=\textwidth]{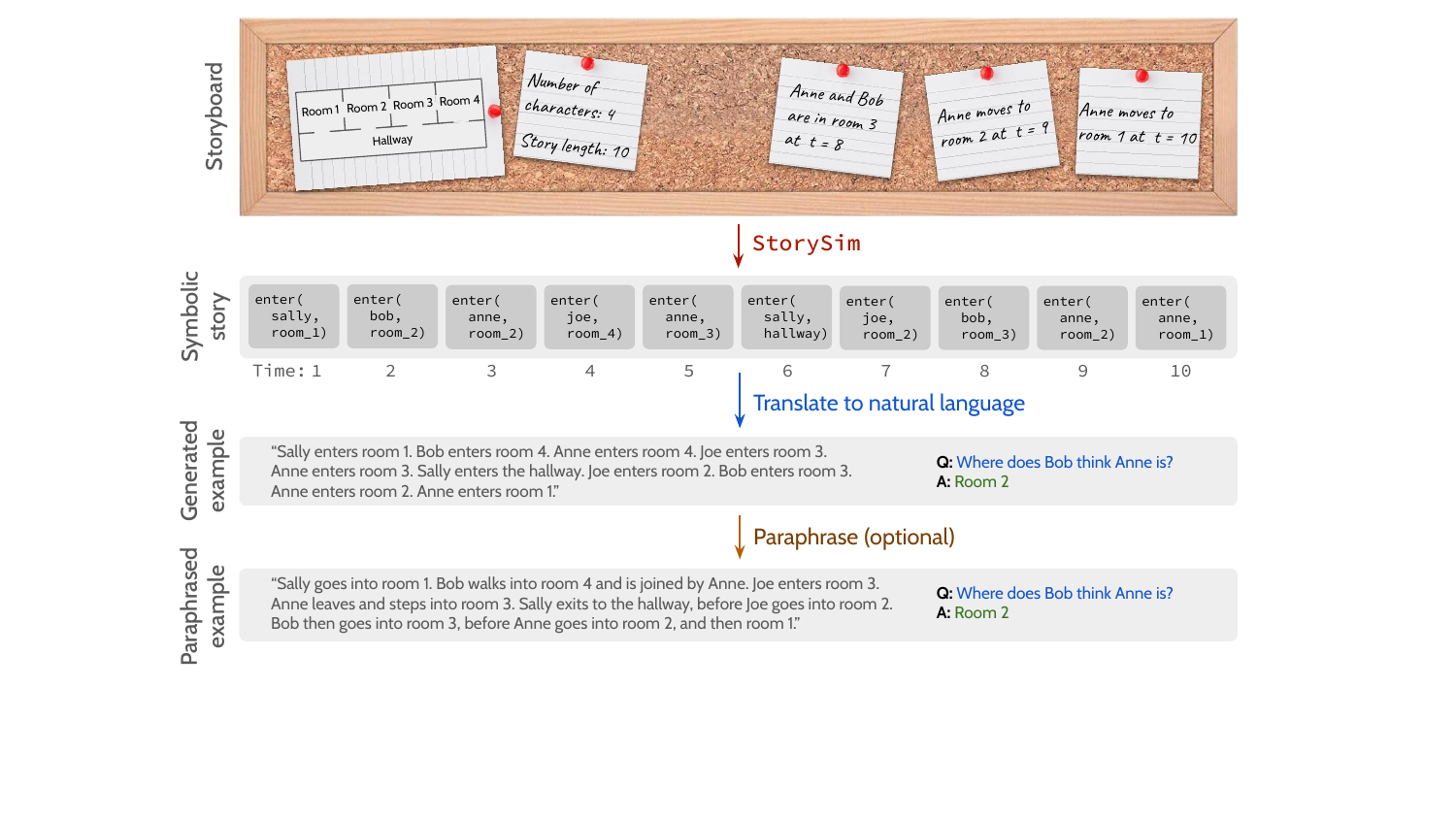}
    \caption{An example \textsc{Storyboard} is shown at the top, consisting of a graph defining the setting, story length, and timestamped key events. \texttt{StorySim} generates a story consistent with the graph and events, where characters move between adjacent locations (starting in the hallway). Characters can observe others’ movements when located in the same room. The resulting events are converted into a natural language QA example and may be further paraphrased into more naturalistic language. This type of story is used in our first-order mislead distance experiments.}
    \label{fig:storysim}
\end{figure*}

LLMs have been evaluated on problems of this nature, and initially showed success \citep{kosinski2023theory}. However, small perturbations of the Sally-Anne test often confound LLMs, indicating that this test has contaminated pretraining data, and LLMs may have overfit to this specific problem, and therefore, they may not necessarily have a sound ToM mechanism \citep{ullman2023large}.

In order to measure ToM ability in a way that is robust against data contamination, we developed a flexible, synthetic benchmark. We propose \texttt{StorySim}, a framework that allows us to generate stories to evaluate ToM and WM abilities of LLMs. The core events of the story are described by a high level specification called the \textsc{Storyboard}. Aside from the \textsc{Storyboard}, the rest of the story is randomly-generated. The generated story is represented symbolically and is guaranteed to be cohesive and consistent. An example of the framework in action is shown in Figure \ref{fig:storysim}. The \textsc{Storyboard} provides a description of the setting in which the story will take place, including which locations are accessible from each other, as well as required actions characters need to make, the length of the story, the number of characters, etc. \texttt{StorySim} then generates a story that is coherent with the \textsc{Storyboard}, filling in all other details randomly in a way that is consistent with the provided constraints. Due to the exact specificity of the events that occur in \texttt{StorySim} via the \textsc{Storyboard}, the randomness of the events in between, and the length of the stories we generate, the chance that any synthetically-generated story in our experiments has appeared in the training data of any LLMs is astronomically small.

\texttt{StorySim} allows us to systematically evaluate current LLMs, which we utilize to address the following research questions:

\begin{enumerate}[itemindent=1.0em,leftmargin=1.6em,noitemsep,topsep=0pt]
    \item[\textbf{RQ1}] Can we measure ToM abilities of LLMs in a way that is robust to data contamination and heuristics?
    \item[\textbf{RQ2}] To what extent do LLMs rely on memorization and heuristics to answer ToM questions?
    \item[\textbf{RQ3}] Can ToM abilities in LLMs be evaluated independently of their general WM capabilities? Relatedly, do LLMs show stronger performance on general WM tasks or on ToM-specific reasoning?
\end{enumerate}

\section{Related Work}


\newcommand{\cmark}{\ding{52}}
\newcommand{\xmark}{\ding{53}}
\newcommand{\neutral}{\ding{108}}

\definecolor{pale_green}{rgb}{0.55,0.75,0.60}
\definecolor{pale_red}{rgb}{1,1,1}
\definecolor{pale_yellow}{rgb}{1, 1, 0.878}

\begin{table*}[t]
    \scriptsize
    \sffamily
    \def\arraystretch{1.15}
    \setlength{\tabcolsep}{2pt}
    \begin{tcolorbox}[
    colback=white,
    colframe=black,
    boxrule=0.5pt,
    boxsep=0pt,
    arc=4pt,        
    left=2pt, right=2pt, top=0pt, bottom=0pt
    ]
        \begin{tabular*}{\columnwidth}{
            m{0.17\columnwidth}|
            >{\centering\arraybackslash}m{0.10\columnwidth}
            >{\centering\arraybackslash}m{0.11\columnwidth}
            >{\centering\arraybackslash}m{0.11\columnwidth}
            >{\centering\arraybackslash}m{0.10\columnwidth}
            >{\centering\arraybackslash}m{0.11\columnwidth}
            >{\centering\arraybackslash}m{0.11\columnwidth}
            >{\centering\arraybackslash}m{0.11\columnwidth}
        }

            \textbf{Benchmark}
            & {\tiny Targets LLM ToM Abilities}
            & {\tiny Guarantees Story Consistency}
            & {\tiny Tests for OOD Generalization}
            & {\tiny Diverse Story Flow}
            & {\tiny Generation Not Dependent on LLM}
            & {\tiny Tests for Higher-order ToM}
            & {\tiny Can Generate Arbitrarily Many Examples} \\
            \hhline{:=:=======:}
    
            \makecell[l]{\texttt{FanTom} \\ \citep{kim2023fantom}}
            & \cellcolor{pale_green!35}\cmark
            & \cellcolor{pale_yellow!35}\neutral
            & \cellcolor{pale_red!35}\xmark
            & \cellcolor{pale_red!35}\xmark
            & \cellcolor{pale_red!35}\xmark
            & \cellcolor{pale_yellow!35}\neutral
            & \cellcolor{pale_yellow!35}\neutral \\
            \hhline{|-|-------|}
    
            \makecell[l]{\texttt{ToMBench} \\ \citep{chen2024tombench}}
            & \cellcolor{pale_green!35}\cmark
            & \cellcolor{pale_red!35}\xmark
            & \cellcolor{pale_red!35}\xmark
            & \cellcolor{pale_red!35}\xmark
            & \cellcolor{pale_red!35}\xmark
            & \cellcolor{pale_red!35}\xmark
            & \cellcolor{pale_red!35}\xmark \\
            \hhline{|-|-------|}
    
            \makecell[l]{\texttt{BigToM} \\ \citep{gandhi2023understanding}}
            & \cellcolor{pale_green!35}\cmark
            & \cellcolor{pale_green!35}\cmark
            & \cellcolor{pale_red!35}\xmark
            & \cellcolor{pale_red!35}\xmark
            & \cellcolor{pale_red!35}\xmark
            & \cellcolor{pale_red!35}\xmark
            & \cellcolor{pale_yellow!35}\neutral \\
            \hhline{|-|-------|}
    
            \makecell[l]{\texttt{OpenToM} \\ \citep{xu2024opentom}}
            & \cellcolor{pale_green!35}\cmark
            & \cellcolor{pale_red!35}\xmark
            & \cellcolor{pale_green!35}\cmark
            & \cellcolor{pale_yellow!35}\neutral
            & \cellcolor{pale_yellow!35}\neutral
            & \cellcolor{pale_red!35}\xmark
            & \cellcolor{pale_red!35}\xmark \\
            \hhline{|-|-------|}
    
            \makecell[l]{\texttt{MindGames} \\ \citep{moore2025large}}
            & \cellcolor{pale_green!35}\cmark
            & \cellcolor{pale_red!35}\xmark
            & \cellcolor{pale_green!35}\cmark
            & \cellcolor{pale_red!35}\xmark
            & \cellcolor{pale_green!35}\cmark
            & \cellcolor{pale_yellow!35}\neutral
            & \cellcolor{pale_red!35}\xmark \\
            \hhline{|-|-------|}
    
            \makecell[l]{\texttt{ExploreToM} \\ \citep{sclar2024explore}}
            & \cellcolor{pale_green!35}\cmark
            & \cellcolor{pale_green!35}\cmark
            & \cellcolor{pale_red!35}\xmark
            & \cellcolor{pale_green!35}\cmark
            & \cellcolor{pale_red!35}\xmark
            & \cellcolor{pale_red!35}\xmark
            & \cellcolor{pale_green!35}\cmark \\
            \hhline{|-|-------|}
    
            \makecell[l]{\texttt{HiToM} \\ \citep{wu2023hi}}
            & \cellcolor{pale_green!35}\cmark
            & \cellcolor{pale_green!35}\cmark
            & \cellcolor{pale_red!35}\xmark
            & \cellcolor{pale_red!35}\xmark
            & \cellcolor{pale_green!35}\cmark
            & \cellcolor{pale_red!35}\xmark
            & \cellcolor{pale_red!35}\xmark \\
            \hhline{|-|-------|}
    
            \makecell[l]{\texttt{ToMi} \\ \citep{le-etal-2019-revisiting}}
            & \cellcolor{pale_green!35}\cmark
            & \cellcolor{pale_green!35}\cmark
            & \cellcolor{pale_red!35}\xmark
            & \cellcolor{pale_red!35}\xmark
            & \cellcolor{pale_green!35}\cmark
            & \cellcolor{pale_red!35}\xmark
            & \cellcolor{pale_red!35}\xmark \\
            \hhline{|-|-------|}
            \hhline{|-|-------|}
    
            \makecell[l]{\texttt{StorySim} \\ (\textbf{ours})}
            & \cellcolor{pale_green!35}\cmark
            & \cellcolor{pale_green!35}\cmark
            & \cellcolor{pale_green!35}\cmark
            & \cellcolor{pale_green!35}\cmark
            & \cellcolor{pale_green!35}\cmark
            & \cellcolor{pale_green!35}\cmark
            & \cellcolor{pale_green!35}\cmark \\
        \end{tabular*}
    \end{tcolorbox}

    \vspace{-0.8em}
    \caption{Comparison of existing ToM benchmarks to ours. \texttt{StorySim} uniquely possesses all of the characteristics described above. Check marks (\cmark) indicate benchmarks that implement the described feature, \xmark's indicate benchmarks that do not, and black dots (\neutral) indicate benchmarks that could feasibly implement the feature, but only with substantial alteration.}
    \label{tab:tom_comparison}
    \vspace{-2.5em}
\end{table*}

\textbf{Theory of Mind}: Existing research benchmarking ToM in language models focuses on prompting with stories followed by question-answering \citep{grant2017can, sap2019socialiqa}, including the \textsc{ToMi} dataset \citep{le-etal-2019-revisiting}. However, these are outdated, and have very likely leaked into the pretraining data of contemporary LLMs. In fact, \citet{ullman2023large} demonstrates that trivial alterations of the \textsc{ToMi} dataset causes model performance to drop. In order to address data leakage, newer benchmarks have been created \citep{wu2023hi, chen2024tombench}. Specifically, we position our work as similar to \citet{sclar2024explore}, where the authors present \textsc{ExploreToM}, a framework that allows for graph-based synthetic story generation and question answering. As in our work, this setup ensures that generated examples are relatively unseen by LLMs. However, while \textsc{ExploreToM} uses an LLM to turn fixed contexts into stories, our framework generates sentences directly from a \textsc{Storyboard}. This gives precise control over when events occur, enabling highly targeted tests (e.g., we can vary the exact location of state-change events in order to probe LLM behavior). In Appendix \ref{sec:adherence}, we demonstrate that LLM-generated stories often fail to meet such constraints, while \texttt{StorySim} reliably produces stories of any specified length, set of characters, and events. Comparisons of our framework and others are further highlighted in Table \ref{tab:tom_comparison}.
\citet{huang-etal-2024-notion2} introduce a complexity measure for ToM tasks and a prompting method that makes implicit character state attributes explicit, offering a formal representation of ToM events similar to ours. However, we test whether LLMs can reason over implicit character states without such scaffolding. Our framework can also add extraneous state changes from irrelevant characters, but as Figure \ref{fig:numppl} demonstrates, these have little effect on ToM performance, contrasting with findings found in previous work.

\textbf{World Modeling}: Though there is much work evaluating aspects of WM abilities of LLMs \citep{kocisky2018narrativeqa, qiu2025phybench, qian2025time}, there are none, to the best of our knowledge, that examine the relationship between WM and ToM ability.  We consider ToM to be an extension of world modeling since ToM requires the explicit modeling of mental states and how they may change over time, and mental states are dependent on the goals, limitations, and perspectives of agents. In the stories we generate, the character's perspectives only consist of what they observe around them, so to reason over this requires understanding the world in which they live and the limitations of each character's perception and knowledge. \citet{wang2024characterbox} train a narrator and a role player model to interact within role-playing game environments. This setup implicitly requires both models to construct and maintain an understanding of the environment (WM), while the narrator must also infer the role player’s perspective to guide narrative choices (ToM). However, the task itself does not directly evaluate either ToM or world modeling capabilities in isolation.

\textbf{Synthetic Datasets}: Though existing synthetic datasets meant for benchmarking LLMs exist \citep{patel2024datadreamer, shvets2025emo, mishra2025kidney}, there are none that evaluate ToM and WM abilities as we do. Synthetic datasets have been shown to be an effective method for training LLMs to perform well on logic tasks \citep{xie2025logic}, clinical text classification \citep{tang2023does}, deductive reasoning \citep{saparov2023testing}, and more.
The diversity of synthetic data is an important factor in the evaluation of performance on both in- and out-of-distribution data \citep{chen2024diversity, havrilla2024surveying}. For this reason, we developed \texttt{StorySim} to be highly programmable and able to generate a wide range of stories.

\section{Methods} \label{sec:Methods}
Below we discuss the details of \texttt{StorySim}'s generation process, the breadth of stories it can create, and how we leverage it to answer our research questions.

\subsection{StorySim} \label{sec:framework}
Broadly, the \texttt{StorySim} framework generates each story in two steps: (1) we construct a \textsc{Storyboard} capturing high-level story specifications, and (2) \texttt{StorySim} uses it to produce a story by generating event sequences that satisfy these constraints while filling in the remaining details coherently.

A \textsc{Storyboard} is defined as $D = (C, \phi, G, E, n)$, where $C$ is a set of characters, $G$ is a directed graph describing the locations in the story, with edges denoting which locations are immediately reachable from other locations, $E$ is a set of programmed events, and $n$ is the number of events. The \texttt{StorySim} framework generates a sequence of $n$ events of the form $\phi(c, l)$, where $\phi$ is the action, $c \in C$ is the character performing the action, and $l \in V(G)$ is the location of the event. Events follow the structure of $G$, so characters move only between adjacent vertices. Only events in $E$ are deterministic, specifying actions at particular time steps; all others are randomly generated but remain consistent with $E$. The exact parameters of $D$ in our experiments are given in Appendix \ref{sec:exactparams}. An example is shown in Figure \ref{fig:storysim}. These events are translated into natural language via templates mapping each $\phi(c,l)$ to a sentence. Optionally, the stories are further paraphrased into more naturalistic language by \texttt{GPT-4o-mini} \citep{hurst2024gpt}.

Though a simple framework, \texttt{StorySim} can be used to create a wide variety of stories. Depending on the desired story, the framework requires varying degrees of manual input. However \texttt{StorySim} can be used to simulate any story where each action can be modeled as a relation (or set of relations) that act over a subject and a node in the graph. The subject does not necessarily need to be a human, and the nodes need not be locations. Additionally, the stories can feature manually-created actions as well, which can add another layer of complexity to the events described by \texttt{StorySim}. Appendix \ref{sec:extrafig} illustrates the range of stories that can be generated by the framework for ToM evaluation, and contains examples of stories we generate in our main experiments, as well as additional stories to further highlight the framework's breadth.

\subsection{Experimental Setup}\label{sec:expsetup}

\texttt{StorySim} can generate stories with either simpler sentences (i.e., template-generated) or more linguistically naturalistic sentences (i.e., paraphrased). In most of our experiments, we utilize the simpler language to minimize confounding effects from linguistic complexity, ensuring errors reflect limitations in the targeted ability rather than misinterpretation---we aim to measure the model's ToM capability independently of the model's ability to parse sentences. In order to explore the effect of linguistic complexity on LLM ToM performance, we run experiments using LLM-paraphrased versions of the stories; these results are reported in Figure \ref{fig:holeswithpara}.

In order to test ToM abilities, we generate two types of ToM problems:

\begin{itemize}[leftmargin=3.0em,itemindent=8.6em,noitemsep,topsep=-4pt]
    \item[\textit{First-order ToM problem}:] Ask about one character's knowledge of another character's location. E.g.: \textbf{``Where does Bob think Alice is?''}
    
    \item[\textit{Second-order ToM problem}:] Ask about one character's knowledge of a second character's knowledge of a \textit{third} character's location. E.g.: \textbf{``Where does Bob think Charlie thinks Alice is?''}
\end{itemize}

To generate stories that allow us to ask questions like the ones above, we create two types of \textsc{Storyboard}s which specify events for characters $S_1$, $S_2$, and $T$, using the Hallways and Doors environment shown in Figure \ref{fig:storysim}. In the above examples, $S_1$ is Bob, $S_2$ is Charlie, and $T$ is Alice. We use \texttt{StorySim} to generate stories for first-order ToM by specifying a high-level pattern within the \textsc{Storyboard}: $S_1$ and $T$ first meet at a shared location, after which $T$ moves elsewhere. Then, without $S_1$’s knowledge, $T$ moves to a third location and remains there for the rest of the story. Similarly, for second-order ToM, we define a pattern where $S_1$, $S_2$, and $T$ initially meet, $S_2$ and $T$ later meet again at a different location, and $T$ subsequently moves to another location alone. Given these \textsc{Storyboard}s, we can ask:

\begin{itemize}[
    before={\let\item\myItem\renewcommand\makelabel[1]{\textbf{##1}}},
    labelsep=0.3em,
    itemindent=0pt,
    listparindent=0pt,
    parsep=0pt,
    partopsep=0pt,
    itemsep=0pt,
    topsep=0pt
]
    \item[\texttt{ToM Prompt}]\label{prompt_A}\hspace{-0.2em}: \textit{Where does $S_1$ think $S_2$ thinks $T$ is?}
    \item[\texttt{WM-Human Prompt}]\label{prompt_B}\hspace{-0.2em}: \textit{When $S_1$ and $S_2$ were in the same room as $T$, where did $T$ go?}
    \item[\texttt{WM-Inanimate Prompt}]\label{prompt_C}\hspace{-0.2em}: \textit{When $S_1$ and $S_2$ were in the same room as $T$, where was $T$ moved to?}
\end{itemize}

The first question type requires the LLM to understand the perspective of $S_1$ and their perceived location of $S_2$ and $T$, thereby directly testing its ToM capabilities. The second question type is equivalent to the first, but instead of asking the model to consider the perspective of $S_1$, we prompt it to algorithmically search through the story to find an event where $S_1$'s perception of the locations of $S_2$ and $T$ changed. Thus, the second question is a test of \emph{world modeling} capability, since we do not ask about the mental state of any character, and answering the question requires understanding the events in the story. Finally, the third question type is equivalent to the second, but instead of describing human subjects, the story describes inanimate objects, where each object is moved by some third-party outside the story. This question is meant to test whether the LLM's world modeling ability is sensitive to whether the entities in the world are animate or inanimate. These experiments enable us to directly test \textit{RQ3}. We must note that the phrasing of the \ref{prompt_C} implies that the objects are moved to different locations by some unnamed force. Our reliance on this unnamed force is unavoidable since objects cannot spontaneously move on their own in the real world, no matter how the story is phrased. The results of evaluating a suite of LLMs on these three prompts are shown in Figures \ref{fig:tomwm}.

Additionally, given the \textsc{Storyboard} for a first-order or second-order ToM problem, we can vary the number of time steps between the event where $S_1$ last perceives $T$'s location and the event where $S_1$'s beliefs are no longer true. A reasonable hypothesis about the ToM abilities of LLMs is that since $T$'s last movement is a more recent event in the story, LLMs may utilize a heuristic where they predict $T$'s last location, despite the fact that $S_1$ did not see this movement. We refer to this number of time steps between these two events as the \emph{mislead distance}. We create \textsc{Storyboard}s for both first- and second- order ToM problems, varying the mislead distance and using the first-order \ref{prompt_A}. We evaluate multiple LLMs on these examples across multiple mislead distances, and conduct an error analysis to see if patterns emerge when LLMs fail to do proper ToM reasoning. The results of this are discussed in Section \ref{sec:mistakes}. Raw accuracies for additional models are presented in Figure \ref{fig:combined_mislead}, and an analysis of these results and as well as an investigation into whether or not models suffer from recency bias are in Appendix \ref{sec:recency}.

\begin{wraptable}{l}{0.5\textwidth}
    \centering
    \renewcommand{\arraystretch}{0.8}
        \begin{tabular}{l|c c}
            \toprule
            
            \textbf{Task} & \textbf{\texttt{GPT-4}} & \textbf{\texttt{R1}} \\
            \midrule
             Sally-Anne   &  100\% &  100\% \\
             Random Story &  75\% &  87\% \\
            \bottomrule
        \end{tabular}
    \caption{{Accuracy on Sally-Anne vs other ToM tasks. (\textbf{top row}) Models are prompted with the exact Sally-Anne setup shown in Section \ref{sec:intro}, in a story with 5 events. (\textbf{bottom row}) Models are given a randomly-generated first-order ToM story of the same length.}}
    \label{tab:sally-anne-results}
\end{wraptable}

We also explore whether and to what extent models have overfit to simpler ToM questions by significantly increasing the number of subjects in the stories. We hypothesize that models may have seen question-answering prompts based on stories during pretraining, but it is unlikely they have seen examples where the stories contain large numbers of characters. Being able to do so would imply that the models can generalize more robustly to stories containing many characters, and would therefore provide evidence of more robust/generalizable ToM and WM capabilities. We prompt the models using the first-order \ref{prompt_A} and \ref{prompt_B} and report the results in Figure \ref{fig:numppl}.

To further demonstrate the breadth of \texttt{StorySim}, we additionally evaluate LLMs on preference attribution and goal-directed ToM tasks, and display their performance in Figure \ref{fig:widetom}. This allows us to contextualize the difficulty of the two mislead distance experiments with other tasks, and gives us insight into which models have the best all-around ToM capabilities. We utilize the same synthetic generation approach to generate examples for these additional tasks as we do in the mislead distance experiments, and therefore are also less likely to be affected by LLM data contamination. Further details about the construction of the goal-directed and preference attribution ToM stories are discussed in Appendix \ref{sec:extrafig}. 

Finally, we repeat the mislead distance experiments using the first-order \ref{prompt_A} under two additional story environments, detailed in Appendix \ref{sec:altenvs}, and compare these results with versions of each story paraphrased by an LLM (\texttt{GPT-4o-mini} These environments are described at length in Appendix \ref{sec:extrafig} To encourage linguistic diversity, we set the temperature to 1.3. This experiment allows us to examine the effects of the templated structure of \texttt{StorySim} stories, as well as the impact of varying relations and location graphs within the \textsc{Storyboard}.

All four of the experiments above present LLMs with stories that are constructed to be dissimilar to stories seen in pre-training data and have answers that can be mapped to certain heuristics, allowing us to answer \textit{RQ1}. Our error analysis in Section \ref{sec:mistakes} allows us to gain an understanding of the heuristics used when LLMs struggle to answer \texttt{StorySim}'s questions, which enables us to test \textit{RQ2}.

\section{Results}

To gauge the difficulty of our benchmark, we asked 12 graduate and undergraduate students, with no prior knowledge of the task, to answer ToM questions about \texttt{StorySim} stories. Each student was shown 50 examples from the Hallways and Doors environment shown in Figure \ref{fig:storysim} across various mislead distances, achieving an average accuracy of 96\%.

\subsection{Sally-Anne Is No Longer a Good Test for ToM}

To highlight the inadequacy of benchmarks that rely on the Sally-Anne problem to measure ToM, we used \texttt{StorySim} to mirror the Sally-Anne experiment using \texttt{GPT-4} \citep{achiam2023gpt} and \texttt{R1} \citep{guo2025deepseek}, using a \textsc{Storyboard} that specifies a first-order ToM problem with 5 events matching the Sally-Anne story setup exactly. As shown in Table~\ref{tab:sally-anne-results}, both of these models had perfect scores on this task, far from the accuracy when prompting the models using randomly-generated stories of the same length. Additionally, in order to get a sense of whether or not our generated stories appear in pre-training data, we performed an n-gram overlap test with the \textsc{C4} dataset, a web crawl corpus often included in pre-training data of LLMs \citep{raffel2020exploring}. We checked for exact matching 5-grams in the first-order mislead distance experiments and their paraphrased equivalents, and found no matching sequences.

\subsection{Evaluating ToM vs WM Capabilities}\label{sec:compare-res}

\begin{figure}[htb!]
    \centering
     \includegraphics[width=0.8\linewidth]{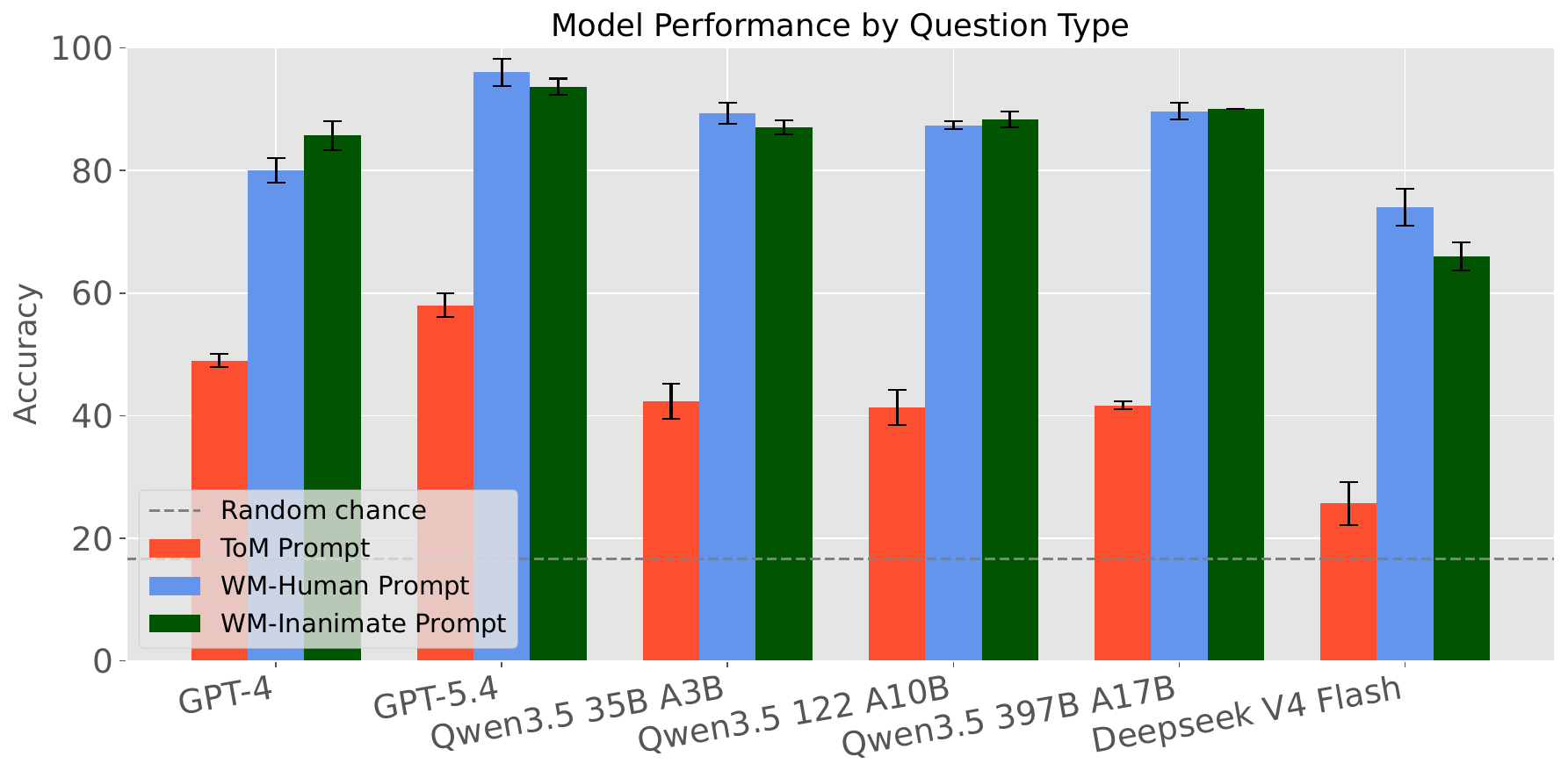}
    \caption{Zero-shot performance per model using the \ref{prompt_A}, \ref{prompt_B}, and \ref{prompt_C}. For each of these prompts, the models are given a story containing 100 events, and we give 100 such stories to each model. Black bars represent the 95\% confidence interval.}
    \label{fig:tomwm}
\end{figure}

We tested a suite of LLMs using the prompts discussed in Section \ref{sec:Methods}. Starting with a \textsc{Storyboard} describing a second-order ToM event, we generate 100 stories consisting of 100 events with which we query each model. 

Figure \ref{fig:tomwm} shows the results of the experiment discussed in Section \ref{sec:Methods} using \texttt{GPT-4}, \texttt{GPT-5.4} \citep{gpt54}, \texttt{Deepseek V4 Flash}, and the \texttt{Qwen 3.5} family of models \citep{qwen35blog}. Across all models, ToM performance is far worse than WM, demonstrating that it is indeed a much more difficult problem. Interestingly, \texttt{GPT-4} and \texttt{GPT-5.4} only have a 9\% difference in performance in the ToM setting despite the two year difference in release dates. The Qwen family of models exhibit nearly identical performance across all 3 settings. In these two families, the models demonstrate a negligible difference in performance between the WM Human and WM Inanimate settings, indicating that modeling the movements of human subjects vs inanimate objects is not significantly different in terms of task complexity. However, \texttt{V4} differs from the others in having the worst performance across the board, and its WM-Inanimate accuracy being much higher than WM-Animate accuracy as compared to all other models.

\begin{wrapfigure}{r}{0.55\linewidth}
    \includegraphics[width=\linewidth]{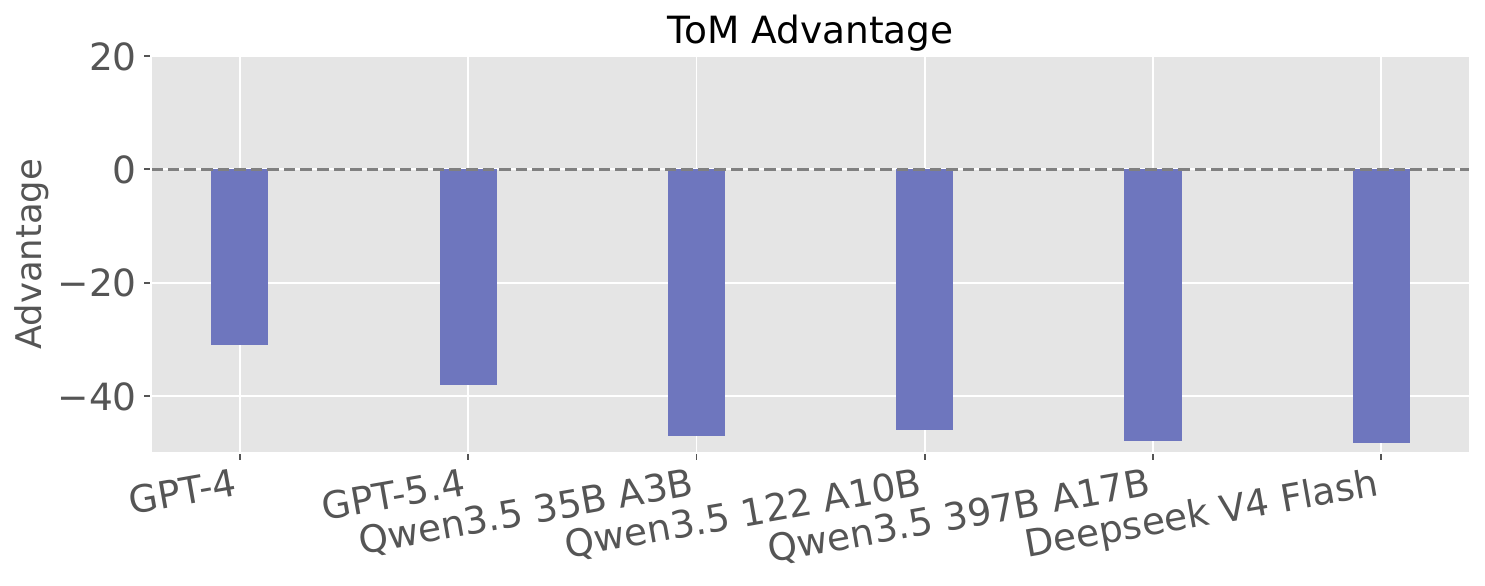}
    \caption{Performance by the \texttt{Llama 3.1 Instruct} family of models, of varying sizes, using 3-shot examples given the \ref{prompt_A}, \ref{prompt_B}, and \ref{prompt_C}. Similar to Figure \ref{fig:tomwm}, these stories have 100 events each.}
    \label{fig:advantage}
\end{wrapfigure}

To understand how ToM performance relates to WM ability, we introduce the \textit{ToM Advantage}: the number of questions answered correctly on ToM but incorrectly on WM, minus the number answered correctly on WM but incorrectly on ToM. A positive value would indicate that a model sometimes succeeds at ToM despite failing the underlying WM task, while a negative value confirms that WM is easier than ToM. As shown in Figure \ref{fig:advantage}, all models have a negative ToM Advantage, confirming that WM is an upper bound on ToM performance. Differences within the \texttt{GPT} and \texttt{Qwen 3.5} families are small, though \texttt{GPT-4} has the least negative advantage of the group. Interestingly, \texttt{V4}'s ToM Advantage is close to the \texttt{Qwen 3.5} models, despite performing far worse on the ToM task. The \texttt{Qwen} models are notable as the only ones that never answer ToM questions correctly while failing the equivalent WM question. Full confusion matrices for each model are provided in Appendix \ref{sec:matrices}.

\subsection{LLM Performance on StorySim ToM Tasks}

\begin{figure}
    \centering
    \includegraphics[width=0.6\linewidth]{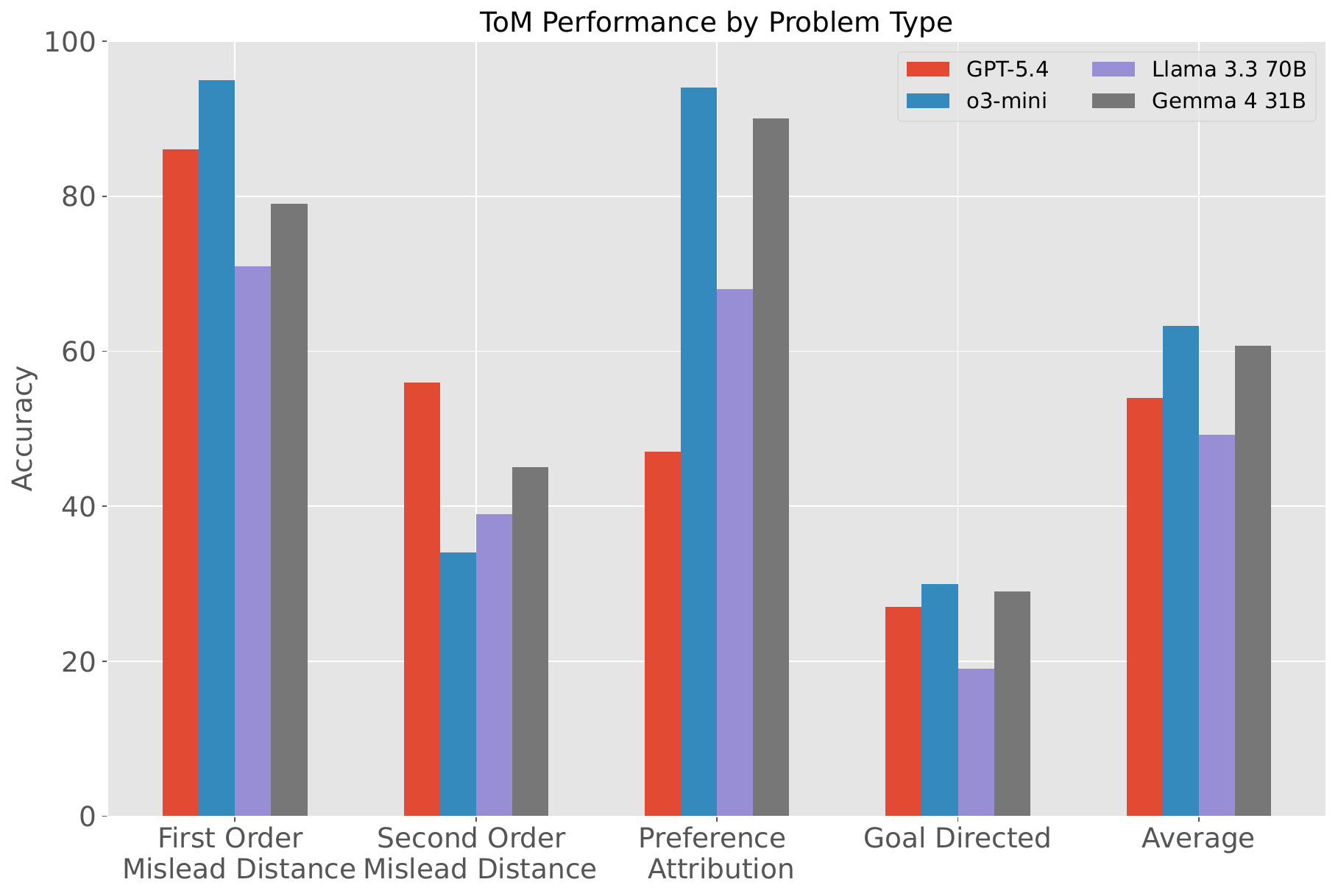}
    \caption{Using \texttt{StorySim}, we evaluate LLMs on the first- and second- order mislead distance task, as well as common other ToM tasks generated by our framework. In the mislead distance experiments, the distance is fixed to 30.}
    \label{fig:widetom}

\end{figure}

Figure \ref{fig:widetom} shows the results of evaluating models on multiple ToM tasks generated by \texttt{StorySim}. Broadly, the first-order mislead distance experiments are the easiest ToM task in our suite of evaluations. The most difficult task seems to be the goal-directed reasoning task, where every model sees its lowest score. Averaging across tasks, \texttt{o3-mini} has the best performance across all tasks, with it being the best performing model in three out of the four tasks. In the second-order mislead setting, however, it has the worst performance across all models. Additionally, the results break the parameter count ordering, with \texttt{Gemma 4 31B} outperforming larger models \texttt{GPT-5.4} and \texttt{Llama 3.3 70B}.

\begin{figure*}[h]
\begin{center}
        \includegraphics[width=0.48\linewidth]{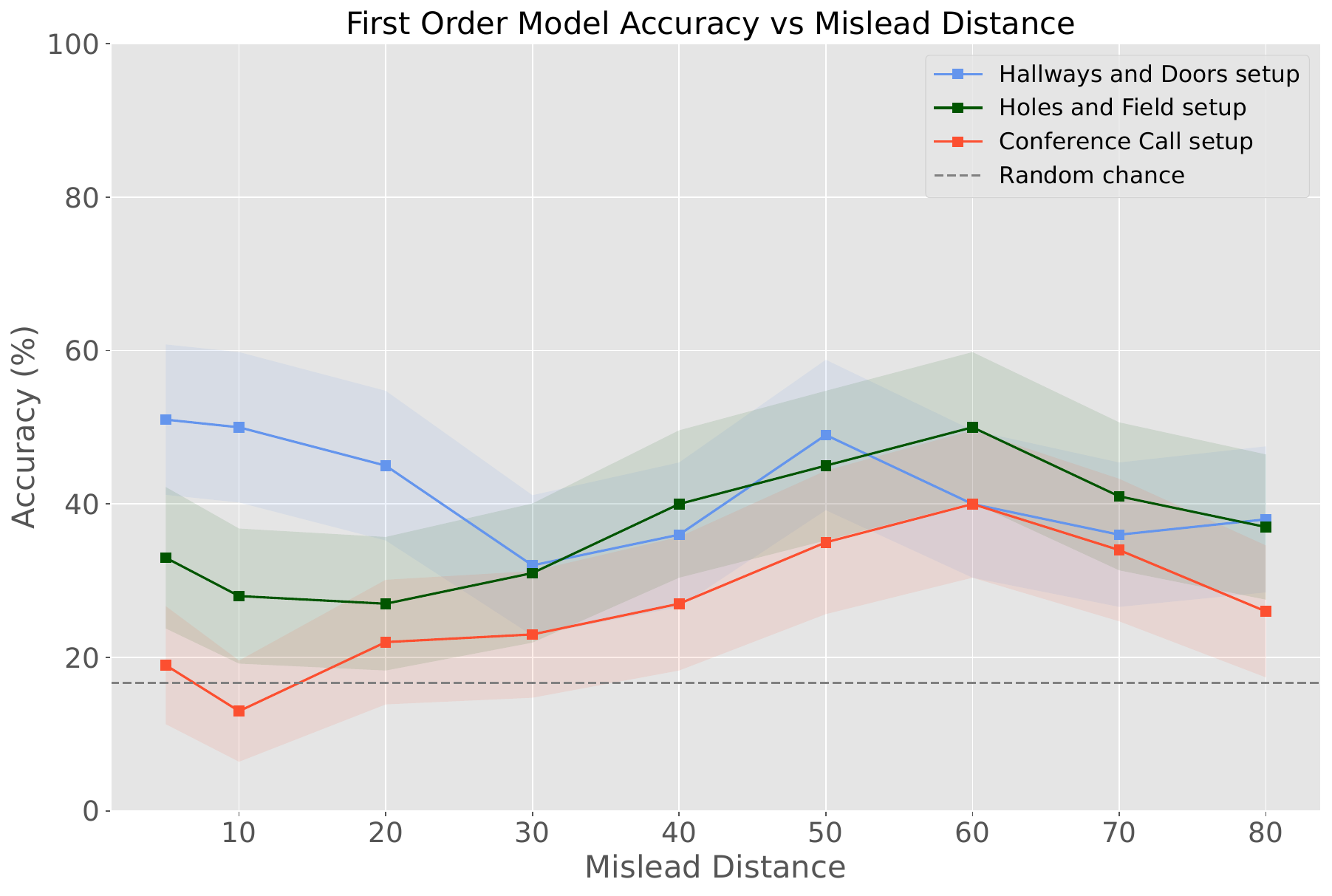}
        \includegraphics[width=0.48\linewidth]{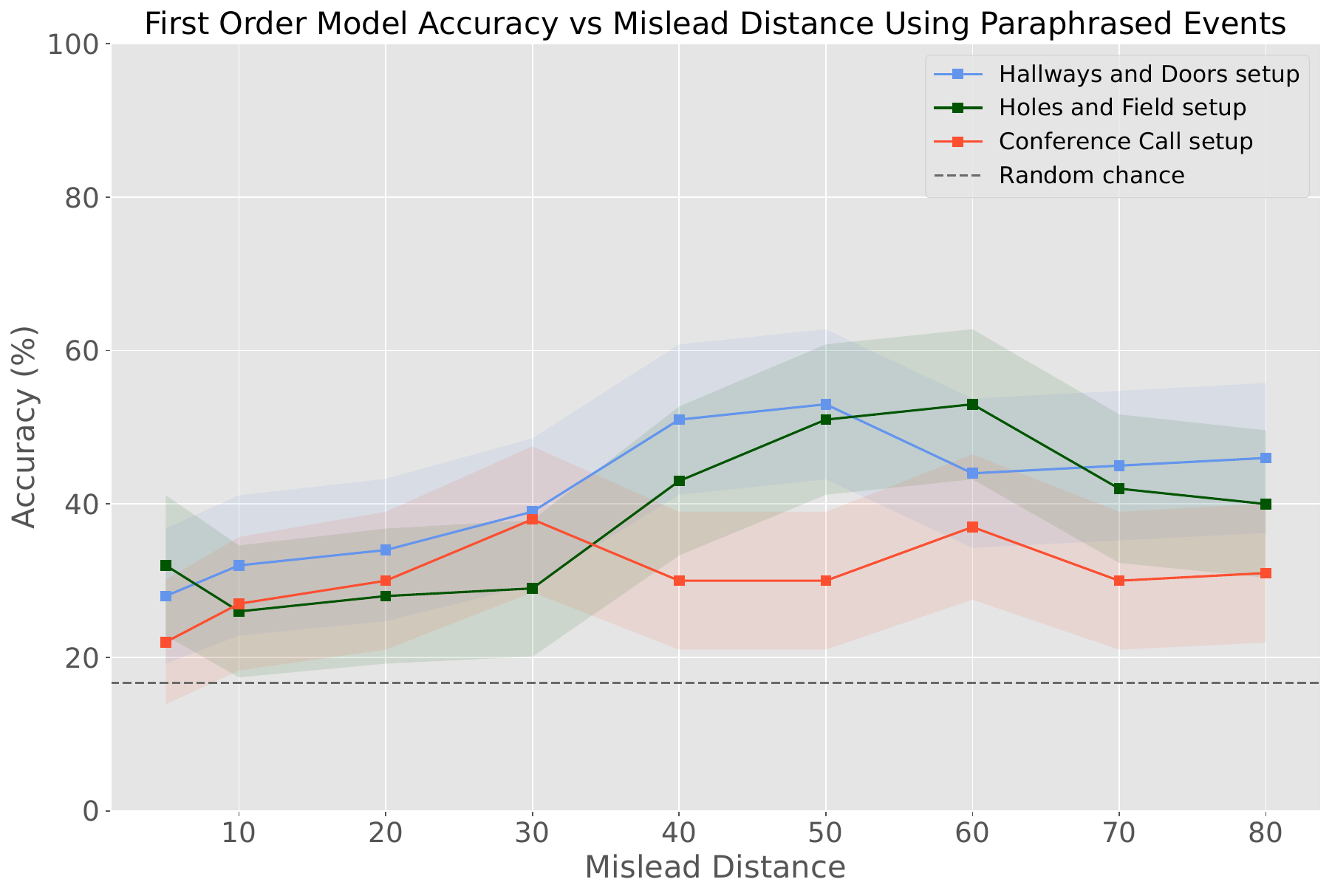}
        \caption{(\textbf{Left}) Model accuracy vs mislead distance using \texttt{GPT-4} using three different \textsc{Storyboard}s featuring characters entering and exiting rooms from a hallway, characters in a field jumping in and out of holes in the ground, and characters joining and leaving conference calls. Otherwise, the stories are generated in the same way. (\textbf{Right}) The same experiment as on the left but using \texttt{GPT-4o-mini} to paraphrase the generated events.}
        \label{fig:holeswithpara}
    \end{center}
\end{figure*}

The mislead distance experiments are constructed with specific \textsc{Storyboard}s as discussed in Section \ref{sec:Methods}. In order to explore the effect of this specific construction on performance, we evaluate LLMs on \textsc{Storyboard}s defined with different $G$ and $\phi$. Figure \ref{fig:holeswithpara} (left) shows that across different environments, performance is similar across most mislead distances. Where larger differences do appear, the Hallways and Doors setup performs best, and therefore we use this setup across all other experiments. As evident in Figure \ref{fig:holeswithpara} (right), paraphrasing the events only causes a meaningful decrease in performance for one story setup at small mislead distances. Otherwise, we do not observe any meaningful difference in performance, suggesting that the simpler language of non-paraphrased examples helps models to solve the task as compared to the more naturalistic language of paraphrased examples.

\subsection{Do LLMs Rely on Heuristics for ToM?}\label{sec:mistakes}


\begin{figure*}[ht!]
    \centering
    \includegraphics[width=0.85\linewidth]{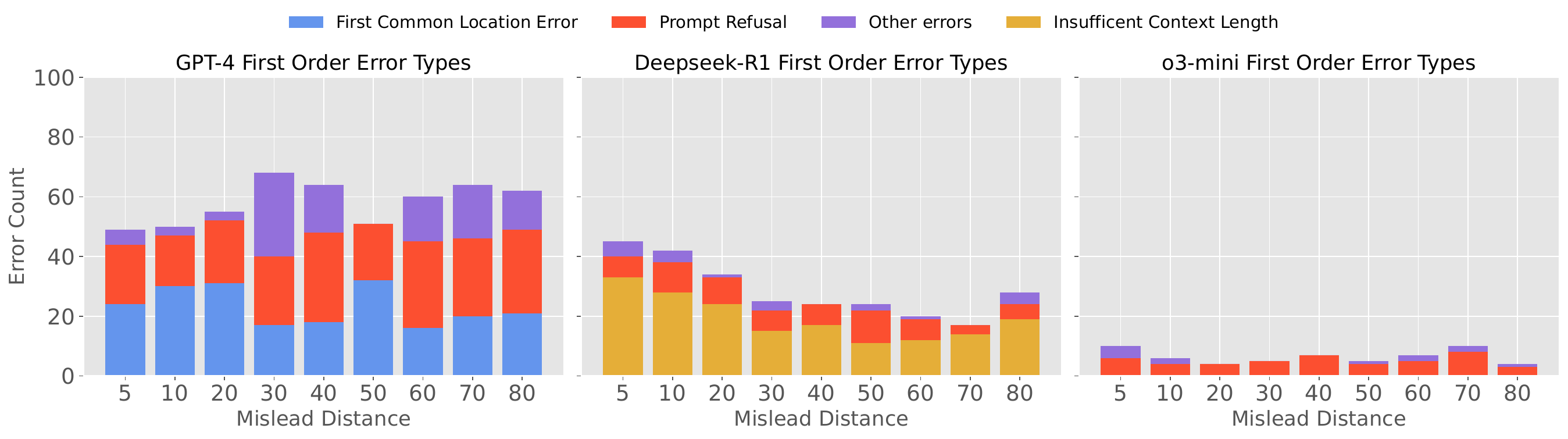}
    \includegraphics[width=0.85\linewidth]{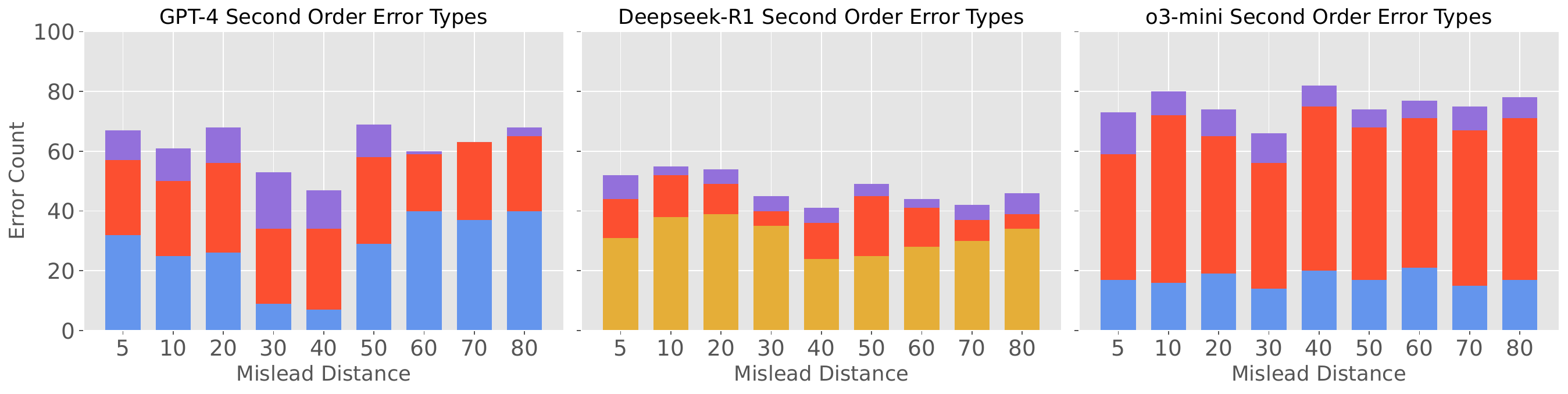}
    \caption{Number of mistaken examples (out of 100), with each example categorized by error type, for the first-order (\textbf{top}) and second-order (\textbf{bottom}) ToM mislead experiments on \textbf{(left)} \texttt{GPT-4}, \textbf{(middle)} \texttt{R1}, and (\textbf{right}) \texttt{o3-mini}.}
    \label{fig:firstordererrors}
\end{figure*}

In this section, we conduct an error analysis to see if LLMs follow specific shortcuts when they fail on our ToM task. In the first-order case, we chose to focus on \texttt{R1} and \texttt{GPT-4} for our error analysis, since the former has the strongest evidence of a performance correlation with mislead distance, and the latter performs the worst overall in the group (see Figure \ref{fig:combined_mislead}). In the second-order case, \texttt{o3-mini} performs the worst despite being the best in the first-order task, so we analyze that as well. The most common error for \texttt{GPT-4} is prompt refusal, where the model responds with ``not enough information'', while for \texttt{R1} it is insufficient context length, where the model runs out of context tokens when generating a long chain-of-thought (see Appendix \ref{sec:collapse} for an example). We attempted to address this by experimenting with different prompts to guide the models, but as demonstrated in Appendix \ref{sec:prompts}, there are no significant differences between the prompt used in the final experiments and other similarly-performing prompts and strategies.

These errors both result in the model not producing an answer, however they have different implications with respect to ToM capability. \texttt{GPT-4}’s prompt refusal indicates that the second-order case elicits a level of complexity that is enough to convince the model that the necessary belief state information is not available. However, as shown in Appendix \ref{sec:collapse}, \texttt{R1} employs the general strategy of reasoning through the events of the story and the evolution of the character’s beliefs. However, it cannot reach the final answer, and instead gets lost trying to re-create the story.

Aside from this, we also found that the most common error is to repeat the location of the first place where characters are in the same location. For example, in Figure \ref{fig:storysim}, this corresponds to predicting \texttt{room\_4}, where Bob and Anne first cross paths en route to \texttt{room\_3}, as specified by the \textsc{Storyboard}. We call this the \textit{First Common Location Heuristic}. For both the first- and second-order ToM tasks, we quantify the relative proportions of each error type in Figure~\ref{fig:firstordererrors}.

In the first-order experiments, \texttt{R1}'s improved performance at larger mislead distances is largely driven by fewer insufficient context length errors. \texttt{GPT-4} exhibits the First Common Location Error at a roughly constant rate across mislead distances, with its accuracy varying mainly due to differing rates of prompt refusal. Moreover, \texttt{GPT-4} predicts locations other than the initial crossing point much more frequently than \texttt{R1}, resulting in worse overall performance.

The second-order experiments do not point to a specific heuristic used by \texttt{R1}; instead, the task appears difficult enough to cause frequent complete failures. \texttt{GPT-4} shows a similar rate of First Common Location errors as in the first-order task. \texttt{o3-mini} performs the worst, and Figure~\ref{fig:firstordererrors} indicates that it commits the First Common Location error far more often than the other models we analyze, despite the fact that \texttt{o3-mini} is specifically trained to perform reasoning tasks and is a relatively new model.

\subsection{Is ToM More Difficult with More Characters?}\label{sec:numchar}

In an effort to determine whether or not LLMs rely on having only seen ToM questions about stories with only a \emph{few} characters, we experimented with increasing the number of characters in the stories. By keeping the length of the story at 100 events, of which only a handful are specified by the storyboard, increasing the number of characters yields stories where an increasing proportion of the events features a novel character.

\begin{wrapfigure}{l}{0.45\textwidth}
    \vspace{-1.6em}
    \begin{center}
        \includegraphics[width=\linewidth]{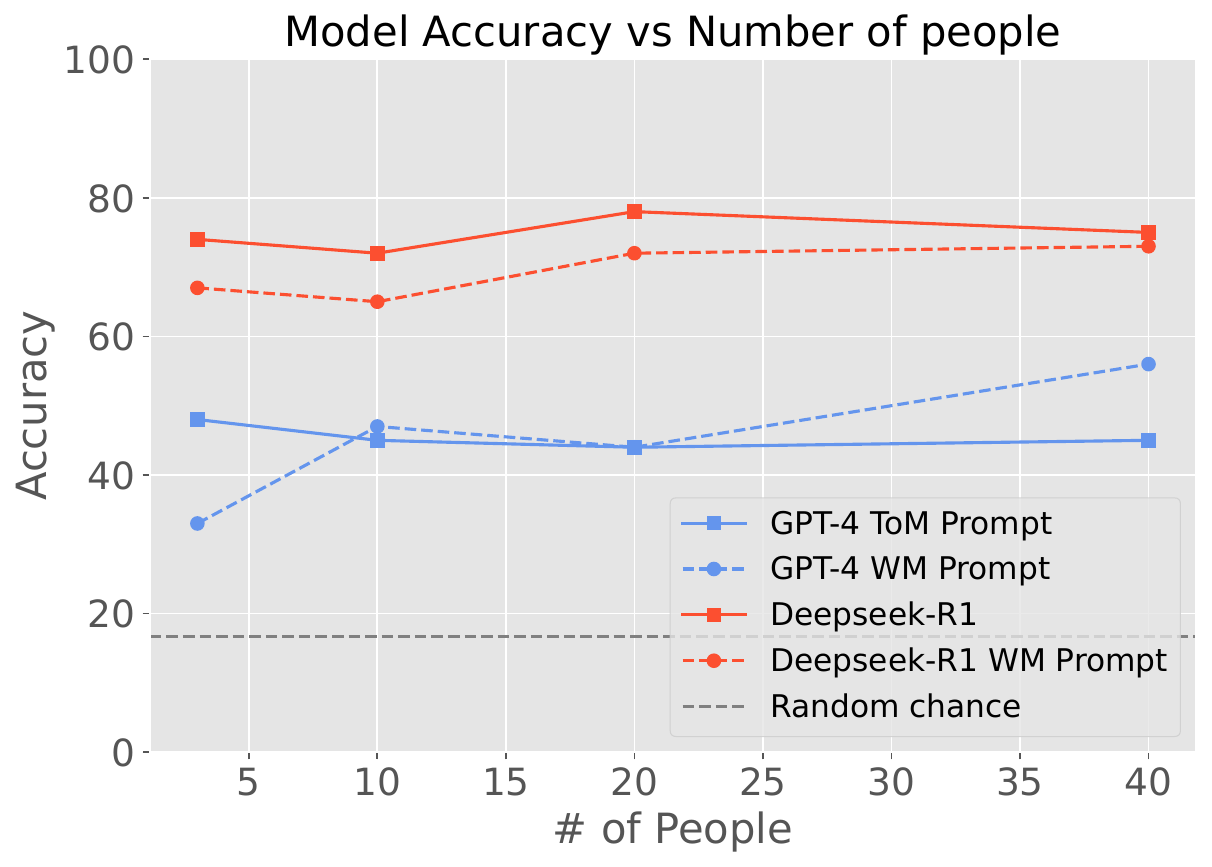}
    \end{center}
    \caption{Model accuracy vs the number of characters in the story using ToM (solid) and WM (dashed) prompts.}
    \label{fig:numppl}
    \vspace{-5em}
\end{wrapfigure}

In the first-order \ref{prompt_A}, the number of characters has negligible effect on ToM performance for either \texttt{R1} or \texttt{GPT-4}. However, when prompted with the WM equivalent of the first-order ToM task, \texttt{GPT-4} struggles with fewer characters but improves markedly as the number of characters increases. This suggests that the presence of more distinct human entities in the prompt elicits more accurate world modeling behavior in \texttt{GPT-4}, enabling it to respond more accurately to our prompts.

\section{Conclusion and Future Work}

We present \texttt{StorySim}, a framework that synthetically generates stories from a \textsc{Storyboard}, a high-level outline containing a number of key events. This framework creates stories that enable us to evaluate the ToM and WM abilities of LLMs in a way that is robust against data contamination, investigate whether world modeling ability is a good predictor of ToM ability, and identify heuristics that LLMs may use in place of ToM reasoning.

Our results show that most LLMs exhibit generally stronger world modeling ability than ToM. Our error analysis reveals that LLMs often memorize incorrect character locations within the stories. Using our framework, we evaluate LLMs on a suite of ToM tasks and find that our goal-directed reasoning task is the most difficult, and that model size does not always predict performance across these tasks.

We hope this work inspires future efforts to benchmark ToM and social reasoning abilities in LLMs in a more controlled and targeted manner. Though we focus on each character's perceptions of other's locations, \texttt{StorySim}'s programmability allows for more complex stories than those presented here. As LLMs are increasingly used in socially-sensitive domains, a rigorous benchmark of their ToM and social reasoning capabilities, as well as the perturbations that affect them, is essential.

\section{Reproducibility Statement}

All generated stories, model outputs, and answer extraction code are made available on \href{https://github.com/ngetachew/StorySim}{\texttt{github.com/ngetachew/StorySim}}. We use the OpenAI Platform API to run experiments using GPT models, and either TogetherAI or OpenRouter for all other models. Unless otherwise stated, all models use temperature 1.0, FP8 precision. Experiments shown in Figure 2 and 3 used models accessed May 31st, 2026 at the latest, except for GPT-5.4, which was accessed March 21st, 2026. All other experiments were run with model versions dated to October 5, 2026. Model outputs are available in a compressed file in the Github repository.




\bibliography{colm2026_conference, anthology, custom}
\bibliographystyle{colm2026_conference}

\appendix

\section{Further Experiment Details}\label{sec:exactparams}

\begin{table}[h!]
\centering
\small
\begin{tabular}{l|p{0.7\linewidth}}
\toprule
\textbf{Component} & \textbf{Specification} \\
\midrule
Characters ($C$) & 8 randomly sampled names \\
Actions ($\phi$) & \texttt{\{enters\}} \\
Location graph ($G$) & 6 vertices (locations) that are fully connected \\
Horizon ($n$) & 100 \\
\midrule
Event sequence ($E$) &
\vspace{-1.2em}
\begin{enumerate}[noitemsep,leftmargin=1.5em]
    \item \texttt{cross\_paths}($S_1$, $T$, location$_1$, $t=10$)
    \item \texttt{move}($T$, second\_location, $t=11$)
    \item \texttt{exclusive\_random}($S_1$, $T$, $t=12 \rightarrow 12+\textit{mislead\_distance}$)
    \item \texttt{move}($T$, third\_location, $t=12+\textit{mislead\_distance}$)
    \item \texttt{exclusive\_random}($S_1$, $T$, $t \rightarrow n$)
\end{enumerate} \\
\bottomrule
\end{tabular}
\caption{Storyboard configuration for mislead distance experiments.}
\label{tab:mislead_setup}
\end{table}

Table \ref{tab:mislead_setup} displays the exact parameters of the mislead distance experiments. The ``exclusive\_random'' event makes \texttt{StorySim} randomly-generate sequences that don't affect the actors specified.
Note that for the second-order experiments, the only difference is the introduction of $S_2$, and an additional cross paths event is added to create the second layer of ToM. In every trial, $S_1$, $S_2$, and $T$ are randomly picked from $C$. Further, $location\_1$, $second\_location$, and $third\_location$ are all randomly generated from $G$. 

The storyboard used for the experiments described in the ToM vs WM capabilities evaluation is the same as the second-order mislead distance experiment, but the mislead distance is fixed at 30. 

\begin{figure*}[htb!]
    \centering
    \includegraphics[width=0.48\linewidth]{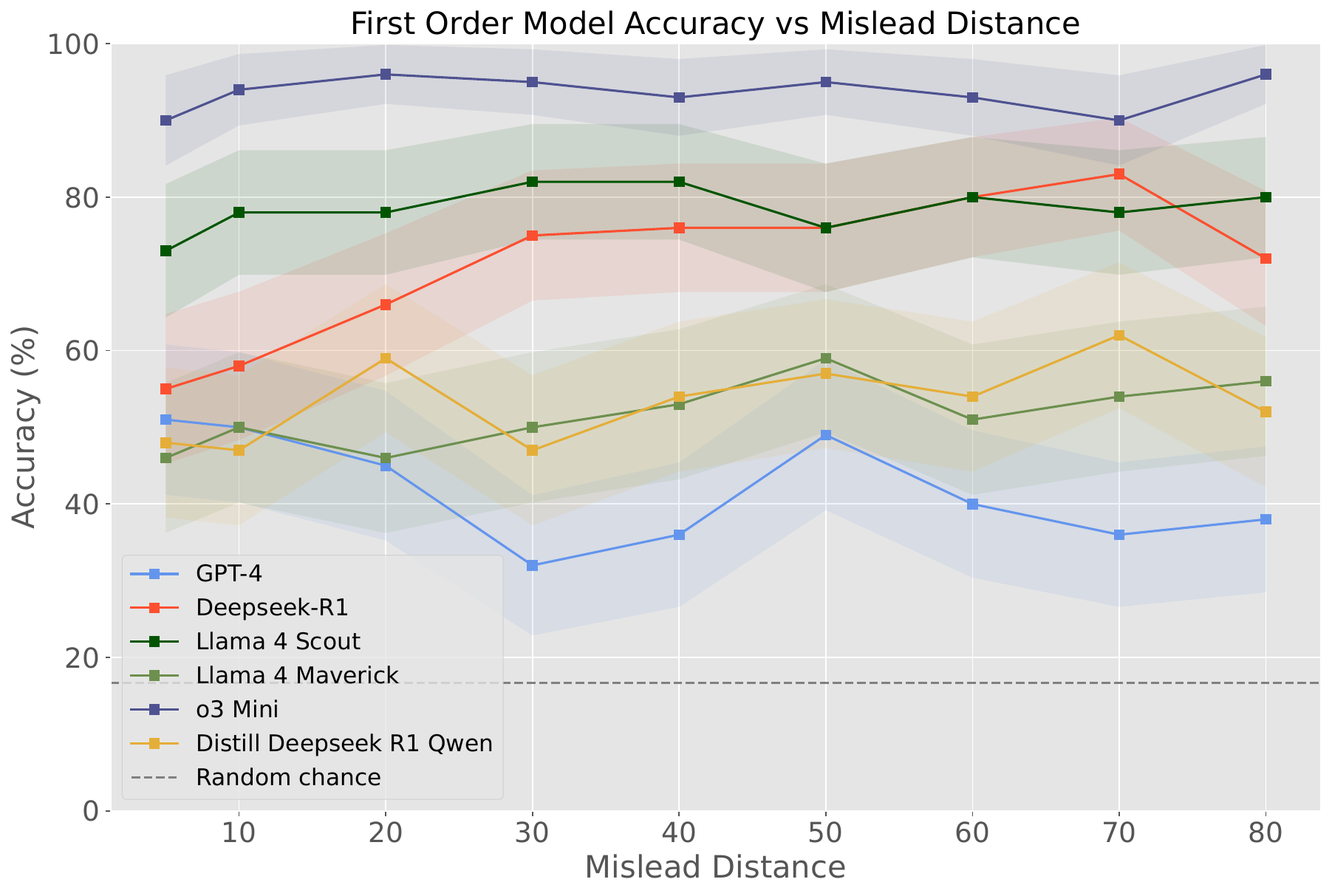}
    \includegraphics[width=0.49\linewidth]{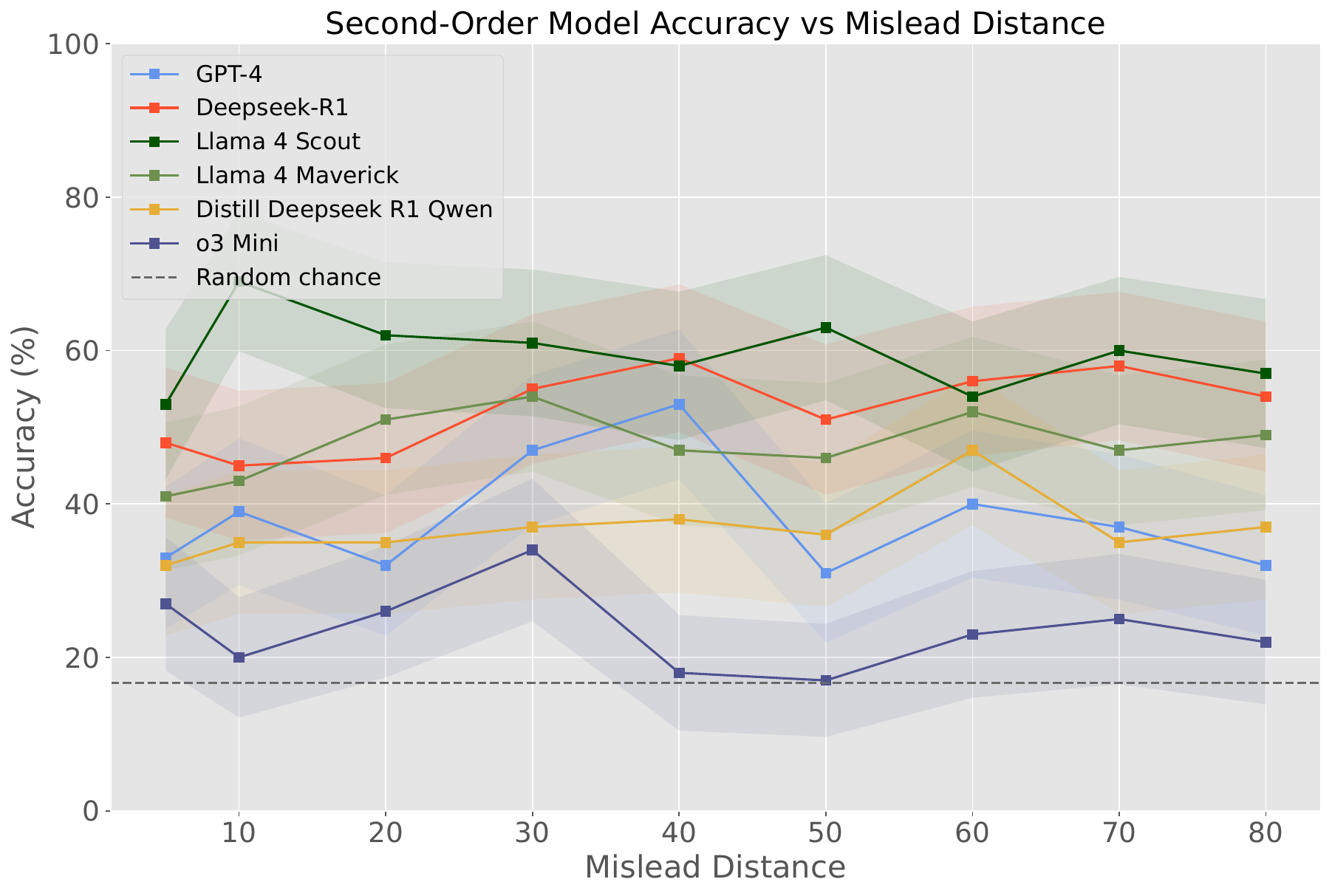}
    \vspace{-0.4em}
    \caption{\textbf{(left)} Model accuracy on the first-order ToM task as a function of the mislead distance (i.e., the number of steps after $S_1$ sees $T$ move to another location and before $T$ moves to a different location altogether). \textbf{(right)} Model accuracy on the second-order ToM task as a function of mislead distance. Shading indicates a 95\% confidence band.}
    \label{fig:combined_mislead}
    \vspace{-0.8em}
\end{figure*}

In the subsequent experiments, we aim to address whether or not LLMs are exploiting simple heuristics to answer our questions rather than relying on an actual ToM mechanism. We use \texttt{StorySim} to generate stories while controlling for numerous confounding variables, including but not limited to the story length, number of characters, locations, events, and the order of events. As a result, the resulting observed differences in LLM behavior can be attributed to specific changes in the experimental variables. 

\section{How Sensitive Are LLMs to Story Construction in ToM?}\label{sec:recency}

A reasonable hypothesis is that LLMs would predict $T$'s last location without considering $S_1$'s mental state, especially if the final location change occurred later in the story. This would be a form of \emph{recency bias}. To test this, we evaluated \texttt{GPT-4}, \texttt{R1}, \texttt{R1-Distilled Qwen}, \texttt{Llama 4 Scout}, \texttt{Llama 4 Maverick}, and \texttt{o3-mini} on first- and second-order ToM problems while increasing the mislead distance (i.e., the number of steps after $S_1$ sees $T$ move to another location before $T$ moves to a different location altogether) \citep{llama4scout, meta2025llama4maverick, deepseek2025distillqwen14b}. We chose this set of models because they can be organized into pairs based on model family with differing numbers of parameters.
Though we see no significant evidence of recency bias, we present values across mislead distances to demonstrate \texttt{StorySim}'s ability to generate targeted stories, and the LLM's variance across slight changes in construction.

In the first-order setting, we see a wide spread of model performances, with \texttt{R1}, \texttt{GPT-4}, \texttt{Qwen}, and \texttt{Maverick} performing nearly identically at smaller mislead distances, then varying more at larger mislead distances. \texttt{Llama 4 Scout} outperforms \texttt{Maverick} despite its smaller size. But in all other pairs of models from the same family, the larger model outperforms the smaller one, providing further evidence of a correlation between model size and ToM ability.

Figure \ref{fig:combined_mislead} (right) shows the results of the second-order ToM version of the experiment. Again, \texttt{Llama 4 Scout} outperforms \texttt{Maverick}, despite the difference in model size, but \texttt{R1} outperforms its distilled \texttt{Qwen} counterpart. The performance gaps between pairs of models of the same family shrinks in the second-order setting, pointing to the difficulty of this ToM task. The most clear differences in performance appear at the largest mislead distances, with the \texttt{Llama} models and \texttt{R1} finishing above the rest. \texttt{GPT-4} is the only model that noticeably improves performance relative to other models. Interestingly, the worst performing model in this setting is \texttt{o3-mini}, which is the best performing model in the first-order setting.

We do not see any correlation between performance and mislead distance, indicating that there is no recency bias at play in the second-order setting. Since \texttt{StorySim} stories are symbolic and relatively simple, we are able to automatically analyze the model's errors in the next section.

\section{World Modeling vs ToM Confusion Matrices}\label{sec:matrices}

\begin{figure}[!htb]
    \centering
    \includegraphics[width=\linewidth]{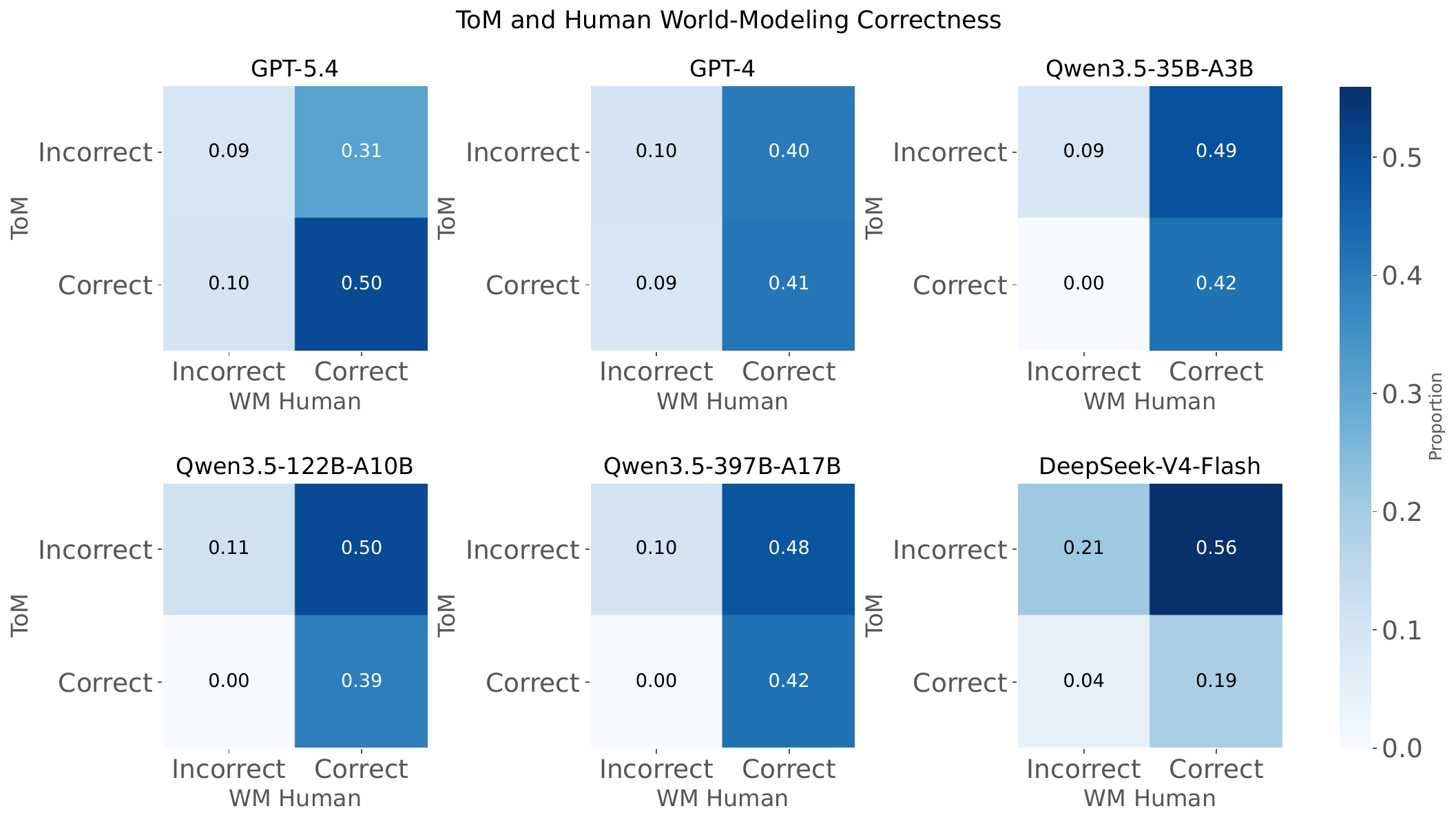}
    \caption{Confusion Matrices for the ToM vs WM Experiment shown in Figure \ref{fig:tomwm}}
    \label{fig:confusionmatrices}
\end{figure}

Figure \ref{fig:confusionmatrices} shows the confusion matrices for the ToM vs WM experiments discussed in Section \ref{sec:compare-res}. Across all models, very few answer story questions correctly in the ToM setting and incorrectly in the WM setting, showing how much more difficult ToM is. That being said, \texttt{GPT-5.4} has the highest amount of this occurring, getting 10 of the 19 examples it answers incorrectly in WM setting correct in the WM setting.

\section{LLM Storyboard Adherence}\label{sec:adherence}

\begin{figure}[!ht]
    \centering
    \includegraphics[width=0.4\linewidth]{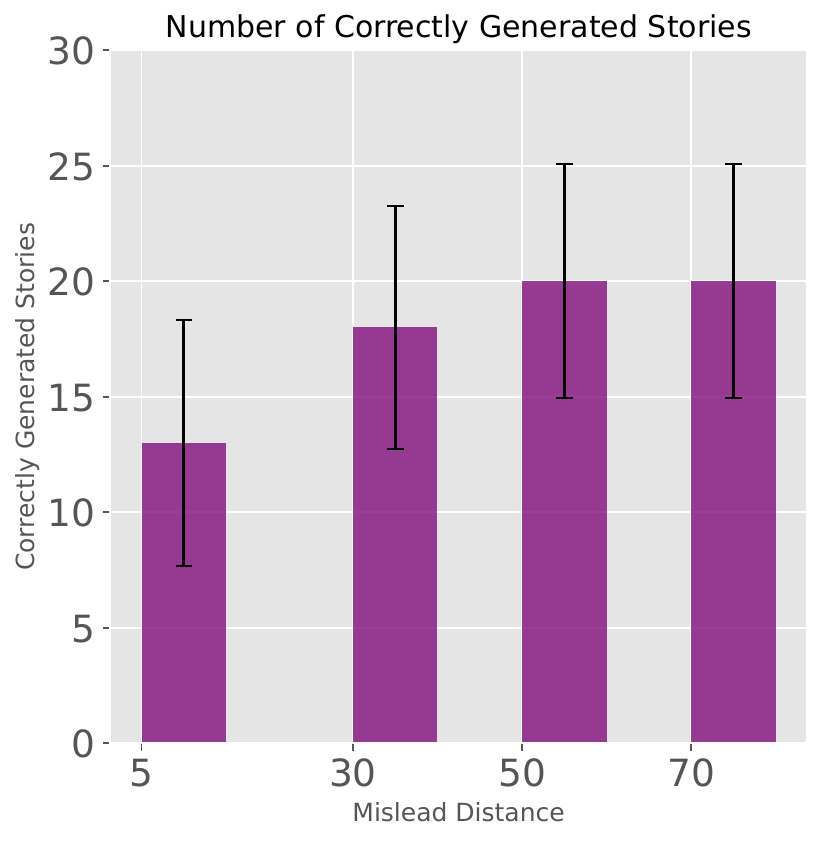}
    \caption{Number of correctly generated stories out of 30 trials using GPT-4. The LLM was prompted to generate stories according to constraints described by a \texttt{Storyboard} that specifies the first-order ToM problem across various mislead distances.}
    \label{fig:llmgeneratedstories}
\end{figure}

 A core feature of \texttt{StorySim} is its ability to generate stories that adhere to the exact specifications of the \texttt{Storyboard}. In Figure \ref{fig:llmgeneratedstories}, we display the performance of GPT-4 generating stories.

\section{Prompt Experiments}\label{sec:prompts}

\begin{figure}[htb!]
    \centering
    \includegraphics[width=\linewidth]{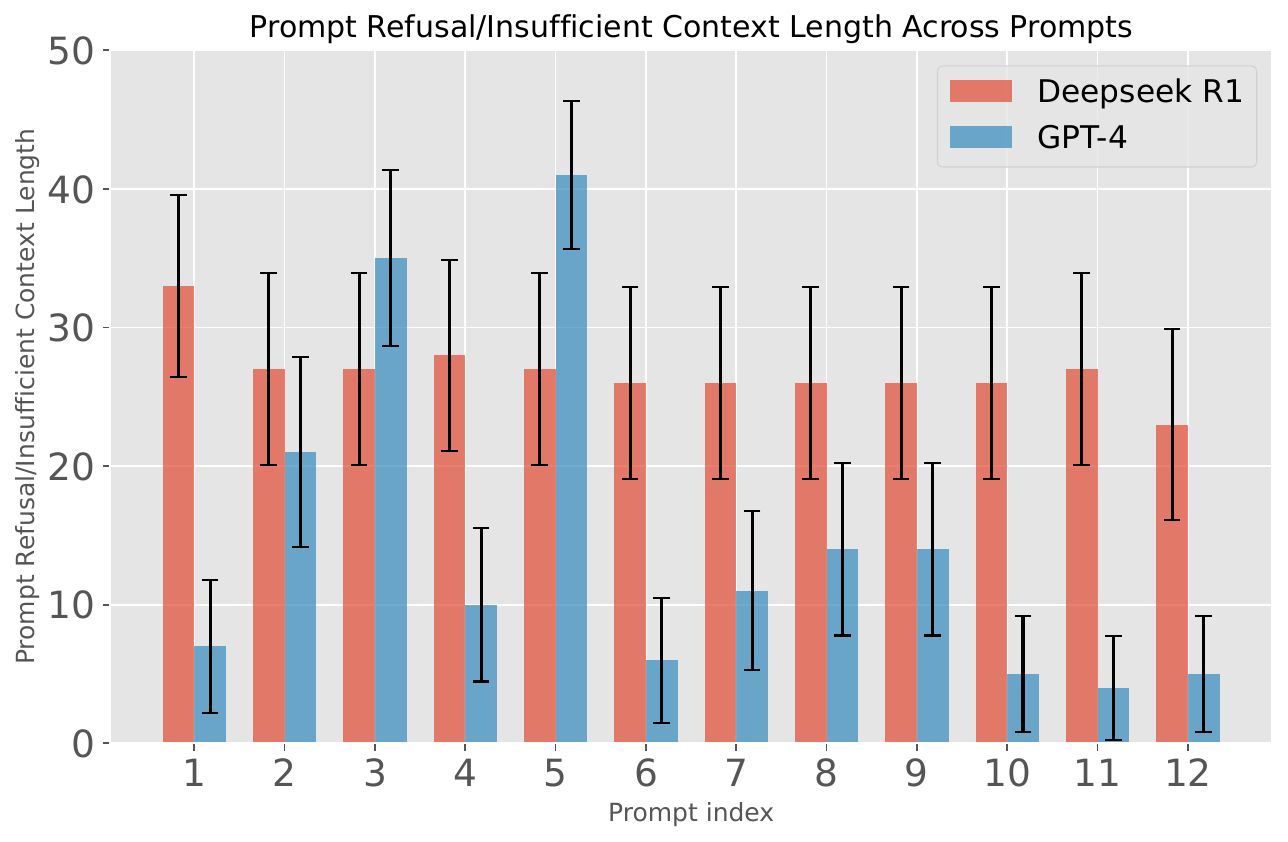}
    \caption{{\rOne Number of prompts refused by GPT-4 and \texttt{R1} over 50 trials, with a fixed \texttt{Storyboard}. Indices on the horizontal axis refer to to prompts shown in Table \ref{tab:full_prompts}. Black bars indicate a 95\% confidence interval. }}
    \label{fig:promptrefusal}
\end{figure}

\begin{figure}
    \centering
    \includegraphics[width=0.7\linewidth]{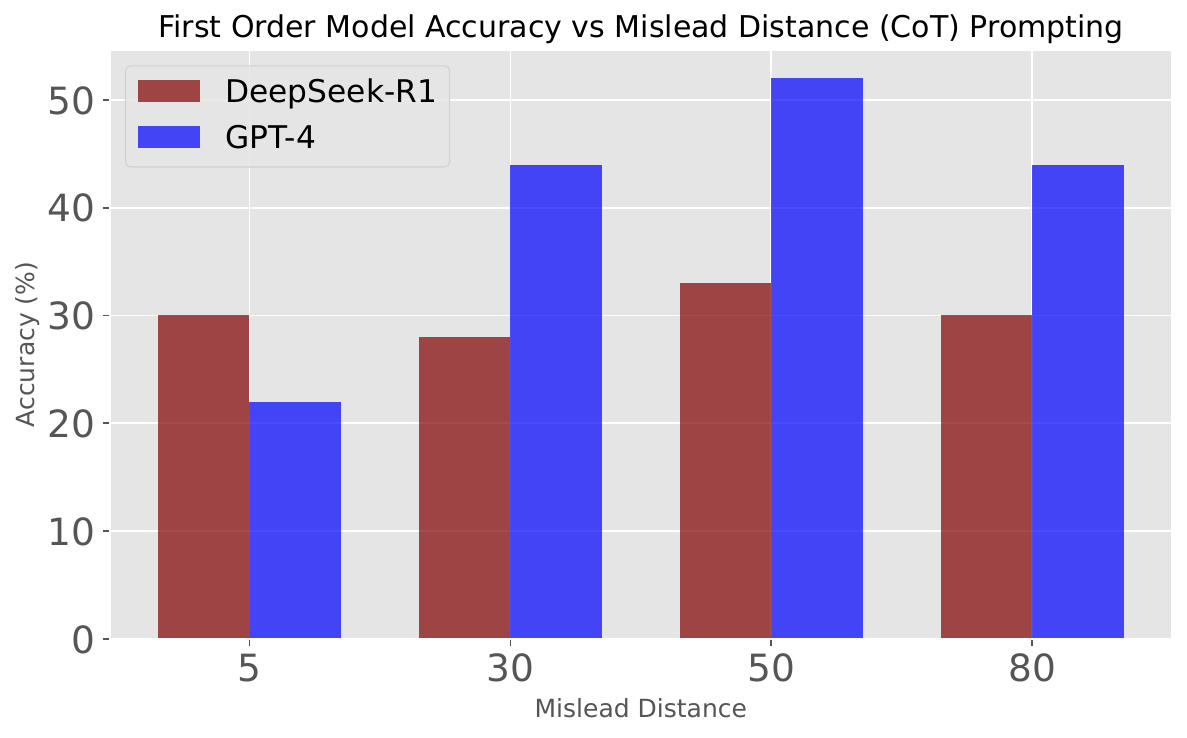}
    \caption{\texttt{GPT-4} and \texttt{R1} performance on the first-order ToM task using chain of thought prompting.}
    \label{fig:cotres}
\end{figure}

\begin{figure}[!ht]
    \centering
    \includegraphics[width=0.7\linewidth]{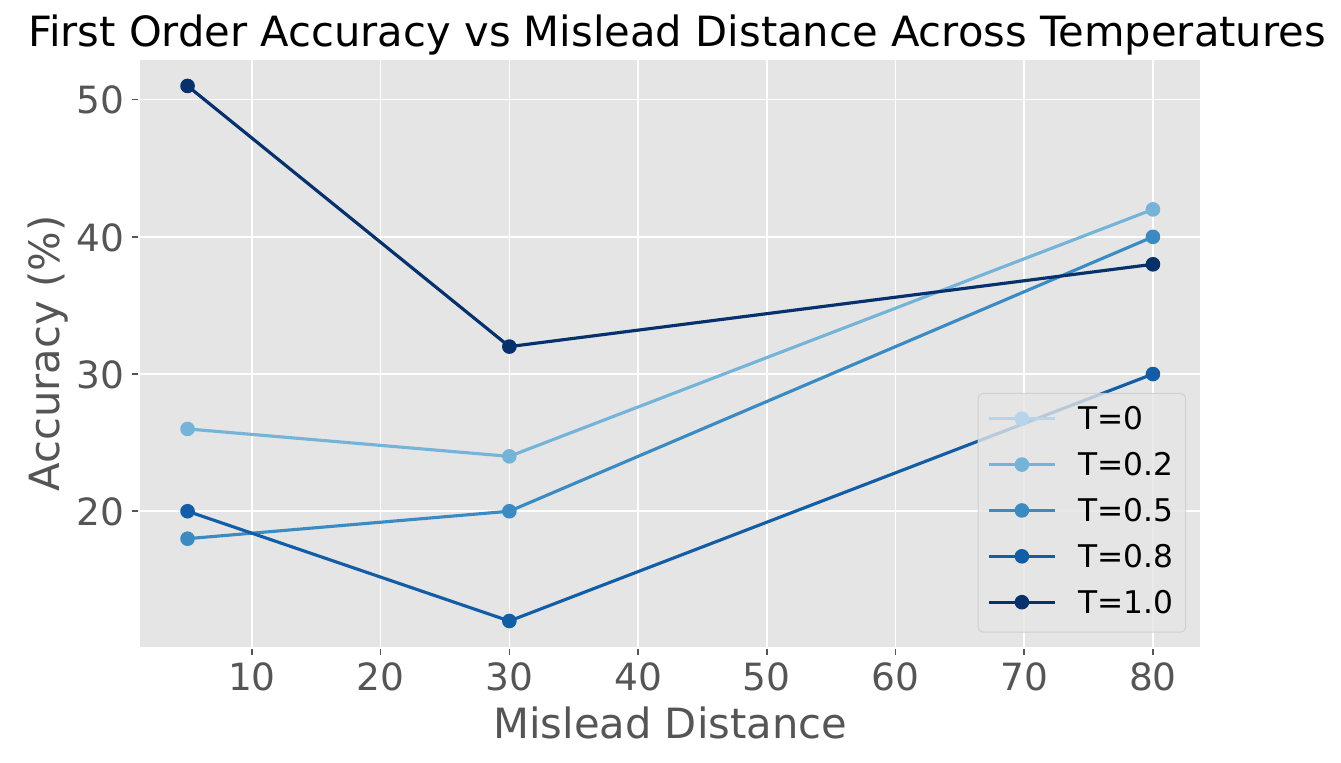}
    \caption{First-order ToM performance of \texttt{GPT-4} using various temperatures.}
    \label{fig:temperatures}
\end{figure}

\begin{figure}
    \centering
    \includegraphics[width=0.5\linewidth]{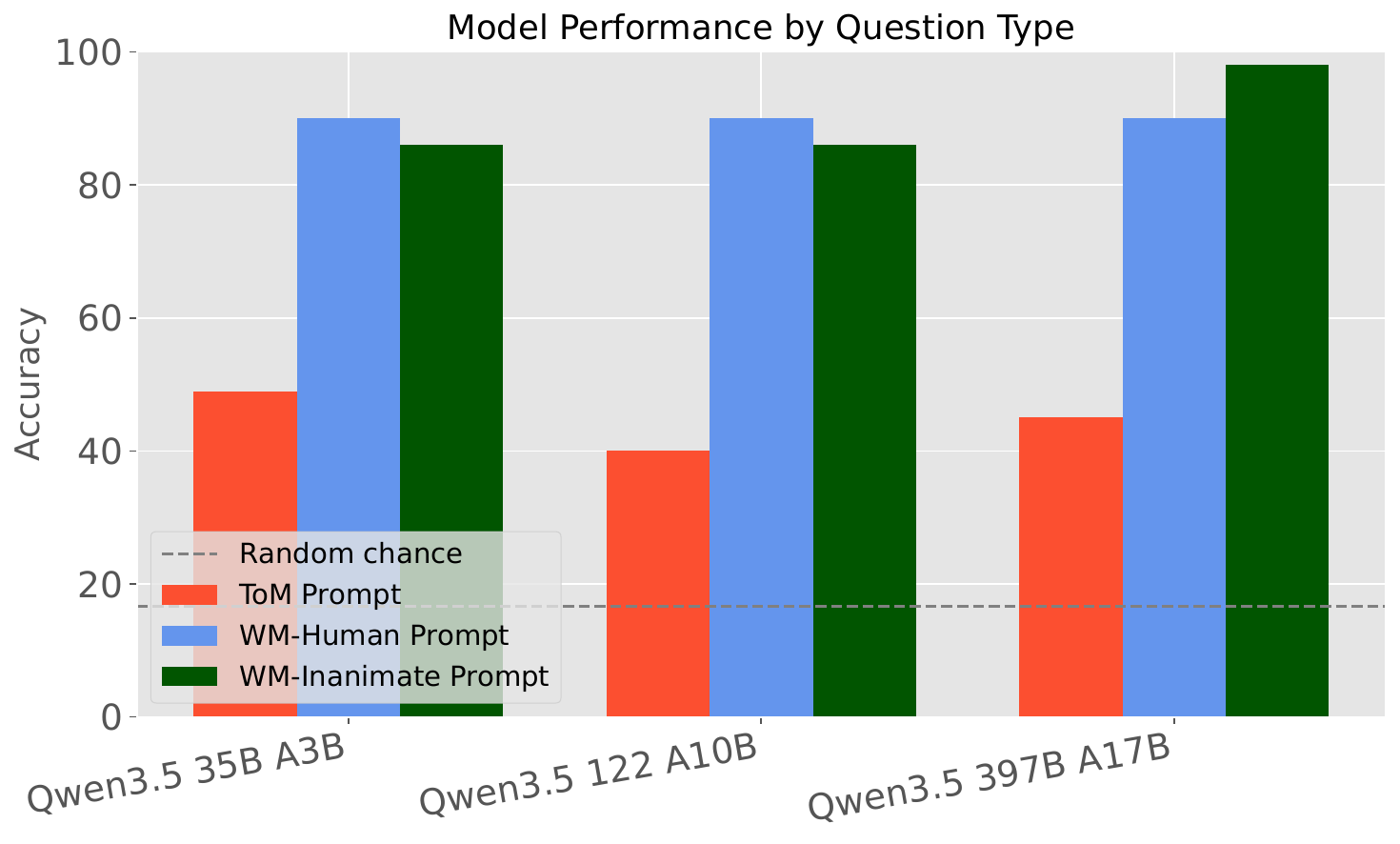}
    \caption{The \texttt{Qwen3.5} family of models performance on the ToM vs WM task given 3 in-context examples each.}
    \label{fig:fewshot}
\end{figure}

In this section, we present supplementary results from experiments with different prompt templates and strategies for avoiding prompt refusal, as initially referenced in Section \ref{sec:mistakes}. We find that there is no significant difference across most of the prompt variants we tested; when large performance differences do occur, they typically reflect that our chosen template (index 1) performs substantially better than the alternatives. Additionally, Figure \ref{fig:cotres} evaluates chain-of-thought prompting with GPT-4 and \texttt{R1} and shows no significant improvement in performance. We further explore the effect of in-context examples across model sizes in the \texttt{Qwen3.5} family in Figure \ref{fig:fewshot}. We see that in the 3-shot setting, the \texttt{Qwen3.5} family performs similar to zero-shot, with the only notable difference being the smallest model, \texttt{Qwen3.5 45B A3B}, gaining 6\% compared to the results in Figure \ref{fig:tomwm}. Finally, Figure \ref{fig:temperatures} reports results for GPT-4 in the first-order ToM setting under varying sampling temperatures. Changing the temperature has either a negligible or negative effect on performance. Overall, these findings suggest that the prompting and sampling configurations used in our main analysis elicit the strongest performance from the evaluated LLMs.

\small
{\onecolumn
    \subsection{Table of Prompts}
    \begin{longtable}{c p{0.85\columnwidth}}
    \toprule
    \textbf{\#} & \textbf{Prompt Text} \\
    \midrule
    
    1 &
    Read the following story and answer the question at the end. Note that all characters start in the hallway. Characters in the same location can see where eachother go when someone leaves. If characters are in different locations, they cannot see eachother. There is enough information to answer every question. \\ \midrule
    
    2 &
    Please read the following story and answer the concluding question. Every character starts in the hallway. Characters who occupy the same location can observe each other's departures, while those in separate locations cannot. All questions are answerable, so be sure to provide an answer. \\ \midrule
    
    3 &
    INSTRUCTIONS: Read the story and answer the question that follows.  
    All characters begin in the hallway. When two characters share a space, they know where the other goes upon leaving; characters in different spaces lack this information. Every question has a sufficient basis in the story, and you must provide an answer.  
    Story: \\ \midrule
    
    4 &
    Review the story and give the answer to the final question. Characters start out in the hallway. Characters in the same room always observe each other's movements, while characters in different rooms remain unaware. The necessary information is guaranteed to be available; provide the answer succinctly. \\ \midrule
    
    5 &
    Read the following narrative and answer the question that appears at the end. All characters begin in the hallway. Shared locations allow characters to see one another move; separated characters have no visibility. Rely solely on the information in the story and give a direct answer. \\ \midrule
    
    6 &
    Carefully read the story and respond to the final question. Every character starts in the hallway. Characters sharing a location know where others go when they leave; characters in different locations do not. The question can be answered from the text alone—state your conclusion briefly. \\ \midrule
    
    7 &
    \textbf{Instructions}  
    Read the story and answer the question that follows.

    \textbf{Rules:} \begin{itemize}
        \item All characters begin in the hallway.
        \item When characters share a location, they observe each other's movements; those in different places cannot.
        \item When characters share a location, they observe each other's movements; those in different places cannot.
        \item The answer is fully determined by the story.
    \end{itemize}   
    
    Provide a concise, final answer. \\ \midrule
    
    8 &
    \textbf{Instructions}  
    You will read a story involving characters moving between locations and then answer comprehension questions.

    \textbf{Rules:} 
    \begin{itemize}
        \item All characters start in the hallway.  
        \item Characters who share a location can track each other's departures; characters who are apart cannot
        \item The question is completely answerable from the story.
        \item Provide a single, final answer.
    \end{itemize} \\ \midrule
    
    9 &
    Instructions: Read the story about characters moving between different locations, then answer the question that follows.

    Rules:
    \begin{itemize}
        \item All characters begin in the hallway.
        \item Characters in the same place can observe each other's departures; characters in different places cannot.
        \item The question is fully answerable from the provided text.
        \item Respond with one clear final answer.
    \end{itemize}
    Story: \\ \midrule
    
    10 &
    You are an assistant tasked with reading a story about characters moving between locations and then answering questions based on it. Follow these rules:
    \begin{itemize}
        \item All characters start in the hallway.  
        \item Characters who share a location can see where others go; characters who are apart lack this visibility.  
        \item The question is fully answerable from the narrative.  
        \item Provide one final answer.  
    \end{itemize}
    If it ever seems like information is missing, re-check the story—it always contains enough clues. Avoid repeating yourself.  
    Story: \\ \midrule
    
    11 &
    You are a helpful reading assistant that will read the provided story about characters moving between locations, then answer comprehension questions at the end. Here are some rules that you must remember:  
    
    \begin{itemize}
        \item All of the characters start in the hallway.  
        \item Characters who share a location know where the other goes when leaving; characters in separate locations have no visibility.  
        \item The question is 100 percent answerable using the given information.  
        \item Output a single final answer.  
    \end{itemize}
    
    If you ever think there isn't enough information in the story, check again, because the questions are designed to be answerable by the prompt. Try not to repeat yourself.  
    Story: \\ \midrule
    
    12 &
    INSTRUCTIONS: Read the following story and answer the question at the end.  
    Note that all characters start in the halllway.  
    If two characters are in the same location, then they know where eachother are.  
    If one of them leaves the location, then the other will know where they went.  
    There is enough information to answer every question. You must provide an answer to every question.  
    
    Story: \\
    
    \bottomrule
    \caption{Full prompt texts used in the experiments.}
    \end{longtable}
    \label{tab:full_prompts}}
\clearpage

\section{Narrative Variations}\label{sec:extrafig}

In this section, we explore different types of stories generated by \texttt{StorySim} some of which are discussed in the main text.
 
 \subsection{Alternative Storyboard Environments}\label{sec:altenvs}
 
 Figure \ref{fig:twoenvs} below shows examples of the stories generated with alternative \textsc{Storyboard} in the experiments discussed at the end of Section \ref{sec:Methods}. Figure \ref{fig:holeswithpara} compares LLM performance on the text shown in Figure \ref{fig:twoenvs} to an LLM paraphrased version of it.

 \begin{figure*}[ht!]
\centering
    \begin{minipage}{0.48\linewidth}
    \begin{tcolorbox}[
colback=gray!15,
colframe=black!70,
arc=3pt,
boxrule=0.6pt,
left=4pt,right=4pt,
top=1.5pt,bottom=1.5pt]
      \small
      \color{RoyalBlue}
      \setlength{\parindent}{1.5em}
      \center
    Georgia joins a call with the city hall.\\
    Bob joins a call with the city hall. \\
    Georgia joins a call with the bank.\\
    ... \\
    Bob joins a call with the doctors office.\\
    \textit{Question}: Who does Bob think Georgia is on the phone with?
    \end{tcolorbox}
    
    \end{minipage}%
    \hspace{1em}
    \begin{minipage}{0.48\linewidth}      
    \begin{tcolorbox}[
colback=gray!15,
colframe=black!70,
arc=3pt,
boxrule=0.6pt,
left=4pt,right=4pt,
top=1.5pt,bottom=1.5pt]
      \small
      \color{RoyalBlue}
      \setlength{\parindent}{1.5em}
      \center
      Frank jumps into hole 1. \\
      Alice jumps into hole 5 \\
      Mike jumps into hole 1. \\
      ... \\
      Mike jumps back out into the field. \\
    \textit{Question}: Where does Frank think Mike is??
    \end{tcolorbox}

    \end{minipage}
    \vspace{-0.7em}
\caption{Examples of two different \textsc{Storyboard} environments used in experiments in the main text. Right is the Holes and Field environment, on the left is the Conference Call environment.}
\label{fig:twoenvs}
\end{figure*}
  
\subsection{Examples of More Complex Stories}

While the generated stories used in our experiments are repetitive in structure, \texttt{StorySim} can generate more complex stories given the correct \textsc{Storyboard}. We provide Figures \ref{fig:threerelations} through \ref{fig:preferenceexample} as a supplementary example of this. If one wishes to make more detailed versions of the stories generated by our framework, it's possible to use an LLM to follow the generated events and generate a polished story. The story shown in Figure \ref{fig:complexstory} is the result of asking \texttt{GPT-4o} to do such a thing.  Figure \ref{fig:goalandcounter} demonstrates two possible stories that can be generated to ask counterfactual and goal-directedness questions that require ToM reasoning. In our main study, we generate the types of stories shown in Figure \ref{fig:goalandcounter} (left) and Figure \ref{fig:preferenceexample}, leaving other types of ToM to future work.

To motivate how \texttt{StorySim} could generate further types of stories, recalling the notation from the Methods section, imagine a story generated where $\phi := whisper$ and $G$ was an adjacency graph specifying a social network of characters who can only whisper to other specified characters. Further, imagine creating multiple \texttt{StorySim} objects and creating a story by sampling randomly generated events from both of them. This could be used to create stories that target characteristics with multiple dimensions. We invite readers to use \texttt{StorySim} to create custom stories that may fit their needs.


\begin{figure}[ht!]
    \centering
    \begin{tcolorbox}[
        colback=gray!15,
        colframe=black!70,
        arc=3pt,
        boxrule=0.6pt,
        left=4pt,right=4pt,
        top=1.5pt,bottom=1.5pt]
            \textcolor{RoyalBlue}{
            \small 
            Charlie arrives at the field. Bob heads to the field. Bob goes to the store. Bob heads to the field Charlie goes to the store. Alice heads to the store. Alice and Charlie exchange phone numbers. Bob goes to the store. Alice goes to the cafe. Bob goes to the field. Charlie arrives at the cafe. Alice calls their friends on the phone to tell them what happened.}
    \end{tcolorbox}
    \vspace{-1em}
    \caption{A story about characters meeting and exchanging phone numbers using the actions \textit{arrives}, \textit{heads}, \textit{goes to}, and \textit{heads to}. Each sentence is a generated event.}
    \label{fig:threerelations}
\end{figure}

\begin{figure*}[ht!]
\centering
    \begin{tcolorbox}[
colback=gray!15,
colframe=black!70,
arc=3pt,
boxrule=0.6pt,
left=4pt,right=4pt,
top=1.5pt,bottom=1.5pt]
    \color{RoyalBlue}
      \small
      \setlength{\parindent}{1.5em}
    Charlie arrived at the field first, scanning the open space with an expectant glance. Moments later, Bob made his way there too, walking at a steady pace. Whatever brief conversation they might have had was short-lived—Bob soon departed for the store.

    His errand complete, Bob returned to the field once more, retracing his earlier steps. Charlie, now done with his own time at the field, headed off toward the store.
    
    Elsewhere, Alice also made her way to the store. It happened that she and Charlie were there at the same time. After a short conversation near the entrance, they exchanged phone numbers before parting ways.
    
    Bob stopped by the store again—not his first visit that day—perhaps tying up a loose end. Meanwhile, Alice took a detour and went to the café. It was quieter there, a place to think or simply pause.
    
    Not long after, Bob returned yet again to the field, following the same path he’d already walked earlier. Charlie, too, arrived at the café—perhaps by coincidence, or perhaps not.
    
    Some time passed. Then, with her phone in hand, Alice made a call. One by one, she reached out to their friends to tell them what had happened.
    \end{tcolorbox}
\vspace{-1.2em}
\caption{Using events from Figure \ref{fig:threerelations} to create a more interesting story by asking an LLM (\texttt{GPT-4o}) to use the structure of the generated events to generate a more naturalistic story, without adding any new events that deviate from the original.}
\label{fig:complexstory}
\end{figure*}

\begin{figure*}[ht!]
\centering
    \begin{minipage}{0.48\linewidth}
    \begin{tcolorbox}[
colback=gray!15,
colframe=black!70,
arc=3pt,
boxrule=0.6pt,
left=4pt,right=4pt,
top=1.5pt,bottom=1.5pt]
      \small
      \color{RoyalBlue}
      \setlength{\parindent}{1.5em}
      \center
    Charlie enters room\_3. \\
    Frank enters room\_2. \\
    Alice enters room\_2. \\
    Danny enters room\_3. \\
    Georgia enters room\_1. \\
    Charlie enters the\_hallway. \\
    Bob enters room\_2. \\
    Frank enters the\_hallway. \\
    Alice enters room\_2. \\
    Danny enters the\_hallway. \\
    Bob enters the\_hallway. \\
    Frank enters room\_4. \\
    Charlie enters room\_5. \\
    Bob enters room\_4. \\
    Alice enters room\_1. \\
    Harvey enters room\_5. \\
    \textit{Question}: Who is Bob clearly trying to follow?
    \end{tcolorbox}
    
    \end{minipage}%
    \hspace{1em}
    \begin{minipage}{0.48\linewidth}      
    \begin{tcolorbox}[
colback=gray!15,
colframe=black!70,
arc=3pt,
boxrule=0.6pt,
left=4pt,right=4pt,
top=1.5pt,bottom=1.5pt]
      \small
      \color{RoyalBlue}
      \setlength{\parindent}{1.5em}
      \center
    Alice enters room\_3. \\
    Charlie enters room\_2. \\
    Jake enters room\_2. \\
    Jake places a football on the ground. \\
    Jake enters room\_4. \\
    Charlie throws the football into room\_1. \\
    Georgia enters room\_5. \\
    Jake enters room\_2. \\
    Jake can't find the ball. \\
    Jake enters the\_hallway. \\
    Jake can't find the ball. \\
    Jake enters room\_1. \\
    Jake found the ball. \\
    \textit{Question}: Where would Jake have found the ball if Charlie didn't throw it in a different room?
    \end{tcolorbox}

    \end{minipage}
    \vspace{-0.7em}
\caption{{\rOne \textbf{(Left)} A story generated to ask a goal-directedness ToM question, as used in the experiments for Figure \ref{fig:widetom}. \textbf{(Right)} A story generated to ask a counterfactual ToM example.}}
\label{fig:goalandcounter}
\end{figure*}

\begin{figure*}[ht!]
\centering

    \begin{tcolorbox}[
colback=gray!15,
colframe=black!70,
arc=3pt,
boxrule=0.6pt,
left=4pt,right=4pt,
top=1.5pt,bottom=1.5pt]
    \small
    \color{RoyalBlue}
      \setlength{\parindent}{1.5em}
      \center
James ponders the expressionist painting. Peter carefully inspects the impressionist painting. Alan checks out the renaissance painting. Victor ponders the expressionist painting. Edward views the fresco. Kevin carefully inspects the expressionist painting. James asks Kevin how anyone could like this painting. Ian ponders the cubist painting. Brian views the minimalist painting. Kevin ponders the expressionist painting. Peter carefully inspects the abstract painting. Ulric ponders the watercolor painting. Jacob ponders the portrait painting. Thomas checks out the expressionist painting. Thomas and Kevin both agree that the piece in front of them is unique. Liam checks out the landscape painting. Victor ponders the oil painting. \\
    \textit{Question}: From Kevin's perspective, who likely doesn't enjoy paintings?
    
\end{tcolorbox}
\vspace{-1.2em}
\caption{A story generated to ask a preference attribution ToM example. This type of setup is use in the preference attribution task shown in \ref{fig:widetom}.}
\label{fig:preferenceexample}
\end{figure*}

\clearpage

\section{Example of Insufficient Context Length Response for R1.}\label{sec:collapse}

Below is an example of one of \texttt{R1}'s responses where it runs out of context length. Note that all errors that it reports finding in the story are untrue. The ellipsis indicates truncation of the prompt for brevity, but where the sentence cuts off in the last step is not truncation. Rather, it is where the model ran out of context length. These erroneous responses from \texttt{R1} often involve the model attempting to enumerate the story’s events, trying to determine when $S_1$ last saw $T$, and then restarting its reasoning chain. When \texttt{R1} does produce a concrete but incorrect answer, it is most often a First Common Location Error, suggesting a tendency to memorize the first location where $S_1$ and $T$ are seen together.

\newpage
\onecolumn
\begin{Verbatim}
<think>
Okay, let's try to figure out where Alice and Bob were last in the
same room and where Alice saw Bob go when he left. 
The question is a bit tricky because I need to track both 
Alice and Bob's movements through all the 
rooms and see when they were last together.
Then, once they part ways, I need to note where Bob
went next from Alice's perspective.

First, I'll start by going through the story step
by step, noting each time Alice or Bob moves. Since all characters
start in the hallway, their initial positions are there. Let's list out their 
movements in order:

1. Edward enters room_1.
2. Alice enters room_1. So now Alice is in room_1 with Edward.
...
35. Georgia enters room_5. Georgia was in room_5, but maybe she left?
Wait, step 23: Georgia was in room_5. Then step 24: Frank leaves room_5.
Then step 25: Charlie enters room_2. Then step 26: Charlie to hallway. 
Step 27: Frank to room_4. Step 28: Danny to room_2. Step 29: Frank to room_3. 
So Georgia was in room_5 until step 35: Georgia enters room_5 again? Wait, 
maybe she was in room_1 before. Let me check. Step 15: Georgia enters room_1.
Then step 23: Georgia enters room_5. So she was in room_1, then room_5.
Then step 35: Georgia enters room_5 again? Wait, maybe that's a typo.
Let me check the original movements again.

Looking back at the original list:

After step 15: Georgia enters room_1.

Then step 23: Georgia enters room_5. So she moves from room_1 to room_5.
...
Step 34: Frank enters room_2.

Step 35: Georgia enters room_5. Wait, but she was already in room_5 at
step 23. So maybe she left room_5 at some point? Let me check. 
The steps after step 23:

After step 23 (Georgia enters room_5), the next movement
involving Georgia is step 35: Georgia enters room_5 again.
But unless she left room_5 in between, she would still be there.
But according to the given movements, characters only move when
specified. So if Georgia entered room_5 at step 23, she stays
there until she moves again. The next move for Georgia is
step 35: Georgia enters room_5 again. Wait, that's
redundant. Maybe it's a mistake, or perhaps she was in
another room. Let me check the steps again.

Wait, step 23: Georgia enters room_5. Then step 24: Frank leaves room_5.
Then step 25: Charlie enters room_2. Step 26: Charlie to hallway. 
Step 27: Frank to room_4. Step 28: Danny to room_2.
Step 29: Frank to room_3. Step 30: Frank to room_4. 
Step 31: Frank to room_3. Step 32: Frank to room_2. 
Step 33: Frank to room_1. Step 34: Frank to room_2.


Georgia's movements:

- Enters room_4 (step 8)
- Enters room_1 (step 15)
- Enters room_5 (step 23)
- Then step 35: Georgia enters room_5 again.
Maybe she was in room_1 again? Wait, no. After step 15, she was in room_1.
Then step 23, she moves to room_5. Then step 35: she's entering room_5 again.
That doesn't make sense unless she left. But according to the given steps,
she didn't leave. So perhaps this is a mistake, 
but I'll proceed as per the given data. 
Maybe she was in room_5 and then enters again, 
which doesn't change her location. So she's still in room_5.

Continuing:

35. Georgia enters room_5. (Still there)
...
53. Danny enters room_1. (From room_4 to room_1)
54. Georgia enters <|EOS|> 
\end{Verbatim}

\end{document}